\documentclass[10pt,twocolumn,letterpaper]{article}

\usepackage{iccv}
\usepackage{times}
\usepackage{epsfig}
\usepackage{graphicx}
\usepackage{amsmath}
\usepackage{amssymb}
\usepackage{booktabs}

\usepackage[breaklinks=true,bookmarks=false]{hyperref}

\iccvfinalcopy % *** Uncomment this line for the final submission

\def\iccvPaperID{4598} % *** Enter the ICCV Paper ID here

\ificcvfinal\fi

\begin{document}

%%%%%%%%% TITLE
\title{Style Generator Inversion for Image Enhancement and Animation}

\author{Aviv Gabbay  \qquad Yedid Hoshen\thanks{Yedid performed this work while at Facebook AI Research.} \\
School of Computer Science and Engineering\\
The Hebrew University of Jerusalem, Israel\\
}

\maketitle
\thispagestyle{empty}

%%%%%%%%% ABSTRACT
\begin{abstract}
One of the main motivations for training high quality image generative models is their potential use as tools for image manipulation. Recently, generative adversarial networks (GANs) have been able to generate images of remarkable quality. Unfortunately, adversarially-trained unconditional generator networks have not been successful as image priors. One of the main requirements for a network to act as a generative image prior, is being able to generate every possible image from the target distribution. Adversarial learning often experiences mode-collapse, which manifests in generators that cannot generate some modes of the target distribution. Another requirement often not satisfied is invertibility i.e. having an efficient way of finding a valid input latent code given a required output image.  

In this work, we show that differently from earlier GANs, the very recently proposed style-generators are quite easy to invert. We use this important observation to propose style generators as general purpose image priors. We show that style generators outperform other GANs as well as Deep Image Prior as priors for image enhancement tasks. The latent space spanned by style-generators satisfies linear identity-pose relations. The latent space linearity, combined with invertibility, allows us to animate still facial images without supervision. Extensive experiments are performed to support the main contributions of this paper. 
\end{abstract}

\section{Introduction}
\label{sec:intro}

Learning image generative models is a fundamental problem for computer vision. A generative image model maps between a canonical probability distribution (which is typically selected to be easy to sample from) and a target image distribution. For every random variable sampled from the probability distribution, the generative model synthesizes an image. The objective for the generator is that the set of randomly generated images will be statistically identical to the true target image distribution. For example, if the target distribution is the set of all faces, the optimal generative model should be able to take as input a random vector and output a realistic looking face. A set of random faces generated by the model, should have an identical distribution to the true distribution of faces. From the objective of generative models, the following properties of a good generative model are implied: i) sufficiency: for each image from the target distribution there should exist an input vector so that when passed through the generator outputs the image ii) compactness: every input vector will generate a realistic image i.e. no generated image will lie outside the target image distribution.

The above properties of generative models make them particularly suitable as image priors. An image prior is a function that scores the likelihood that a given image comes from the target image distribution. Given a generative model of the target distribution, if an image can be generated exactly by some vector input to the generator, then the image comes from the target distribution. The opposite should also be true, if the image cannot be generated by the model, then it does not come from the distribution. Even if perfect generative models are not in fact available, it is assumed that images coming from the target distribution should be possible to be generated with a small error.

At present, the state-of-the-art neural networks for training high-resolution image generators use adversarial training \cite{dcgan, infogan, arjovsky2017wasserstein, karras2017progressive, brock2018large}, which is a powerful but tricky technique. Although adversarially-trained generators have been able to generate very realistic looking images, they suffer from mode-dropping i.e. many valid images cannot be generated using the models. This violates the requirement that for every target image, there exists a valid input which will generate it. Another related issue is non-invertibility: inability to efficiently find a valid latent code that reconstructs a given image. This can occur even for fully sufficient generators, if the optimization of generator inversion is computationally hard. Lack of invertibility prevents SOTA GAN-trained generators from being used as strong image priors.

Very recently Karras et al. \cite{karras2018style} introduced StyleGAN, a new generator architecture that breaks away from previous adversarial generator architectures by re-factoring the latent code as per-channel factors and biases that are applied directly to all layers. In previous GANs, latent code information was only introduced at the bottom layer. In this paper, we present the surprising result, that style-generator are invertible even when trained using adversarial training.

The invertibility of style-generators has several important consequences: Dramatically breaking from traditional GAN-trained generators, StyleGAN-trained models can be used as high-quality image priors. In extensive experiments, we show that the StyleGAN-based priors can help achieve success in challenging image restoration tasks and significantly outperform previous GAN generators as well as the generic Deep Image Prior.

An additional advantage of invertibility is being able to use the linearity of semantic attributes in the latent code space spanned by adversarially-trained generators. Specifically, we use the linearity and independence of identity and pose latent codes, to transfer the pose from a video of one person to a still image of another person. We are able to apply this technique to achieve unsupervised face reanimation at $1024 \times 1024$ facial resolution. The technique is training-free and can be applied in a plug-and-play fashion to new poses and identities. We present experimental evidence, illustrating the effectiveness of our novel method.

Our paper makes the following contributions: i) Recognition of the invertibility of style-based generator architectures. ii) Introduction of a state-of-the-art image domain prior. iii) An unsupervised technique for still image reanimation. Code for the different techniques presented in this paper is available on our project page\footnote{http://www.vision.huji.ac.il/style-image-prior}.

\section{Previous Work}
\label{sec:prev}

\textbf{Generative Modeling:} Generative image modeling has attracted research from the community for several decades. An example of an early approach is using Mixtures of Gaussian models (GMMs) for modeling natural images \cite{zoran2011learning}. Like other early approaches, the images generated by GMMs are not of very high resolution or quality. 

The advent of deep learning revolutionized generative image models. Current deep learning-trained generative models typically differ in their compactness and sufficiency properties. Some are compact (every generated image comes from target distribution) and other models are sufficient (all target distribution images can be generated by the trained model). Until now, models have not been able to satisfy both properties even approximately.

 \textbf{Non-Adversarial Generative Models:} Variational Autoencoders (VAEs) \cite{vae}, introduced by Kingma and Welling, presented a strong non-adversarial method for training generative models. Generators trained by VAEs are typically invertible but not compact (generations are blurry and with missing details), they typically also do not scale up to high-resolution. Consequently, VAEs do not make for particularly good image priors. Followups include the Wasserstein Autoencoder \cite{tolstikhin2017wasserstein}, which has similar properties to vanilla VAEs.
 
 There are several other methods which are easy to invert including: Generative invertible flows \cite{dinh2014nice}, and their later extention to higher resolution  \cite{kingma2018glow}. Such methods are however prohibitively computationally expensive, and do not achieve comparable quality to the most recent GANs. 

\textbf{Adversarial Generative Models:} The state-of-the-art in generative image models is achieved by Generative Adversarial Networks (GANs), introduced by Goodfellow et al. \cite{goodfellow2014generative}. GANs train image generators by joint training with image discriminators. At the end of training, the distribution of images synthesized by the generator is expected to be so close to the target distribution, that no discriminator can distinguish between them. Differently from the above, GAN models are compact i.e. nearly every generated image comes from the target distribution (images look realistic). GANs however are not sufficient, they suffer from mode dropping and can therefore only generate some parts of the target distribution. GANs are therefore not invertible. Many methods were proposed for improving GANs e.g. changing the loss function (e.g. Wasserstein GAN \cite{arjovsky2017wasserstein}) or regularizing the discriminator to be Lipschitz by: clipping \cite{arjovsky2017wasserstein}, gradient regularization \cite{gulrajani2017improved, Mescheder2018ICML} or spectral normalization \cite{miyato2018spectral}. GAN training was shown to scale to high resolutions \cite{zhang2018self} using engineering tricks and careful hyper-parameter tuning. In this paper, we show that StyleGAN \cite{karras2018style} break away from previous architectures, by satisfying the sufficiency property.

\textbf{Image Priors:} Image processing often deals with poorly specified problems that require assigning scores to alternative feasible solutions. Classical approaches for selecting the most natural images include: smoothness and sparsity. More recently, deep learning approaches have significantly improved the quality of image priors. A popular prior model is an adversarial loss \cite{wang2016generative, pix2pix}, which determines if an image comes from the target distribution. This loss function requires per-task training, which is disadvantageous if many tasks are attempted, tasks are not known apriori or that tasks are data poor. The Deep Image Prior (DIP) \cite{ulyanov2018deep} was recently introduced as a generic image prior. It was found to perform well on different image processing techniques (denoising, debluring, super-resolution). It does not however encode any domain knowledge, and therefore fails whenever such knowledge is required. GAN-trained generators have been proposed \cite{zhu2016generative} as domain specific image priors. The lack of invertibility however significantly reduced their effectiveness (as they often could not model test or even train images). In this work we show that style-based architectures overcome those issues.

\textbf{Animating Still Images:} Still image animation has been an active research topic for decades \cite{blanz1999morphable, breuer2008automatic}. Recently, it has been shown that with fiducial point supervision \cite{elor2017bringingPortraits, wang2018video}, facial pose can be transferred between a video and a still photograph. In this work, we show that using generator invertibility and the linearity of the latent space with respect to identity and pose, unsupervised high-resolution reanimation can be performed.  

\section{Method}
\label{sec:method}

In this section, we will describe a method for creating invertible representations. We will then describe how the generative model can be used as an image prior. We will also describe another application: animating still images.

\subsection{StyleGAN}
\label{sec:method:stylegan}

\iffalse

\begin{figure}
\begin{center}
\includegraphics[width=0.7\linewidth]{chart/mapping.jpg}
\end{center}
\caption{An illustration of the mapping network which takes as input noise vector $s$ and returns latent code $z$}
\label{fig:mapping}
\end{figure}

\begin{figure}
\begin{center}
\includegraphics[width=\linewidth]{chart/synthesis.jpg}
\end{center}
\caption{An illustration of the synthesis networ which taken in a set of latent codes $z_0..z_L$ and returns an image G(z). The latent codes modify the activations of the convolutional layers as described in Fig.~\ref{fig:adain}. During generator training all $z_l$ have the same value, during latent optimization we let the latent codes $z_l$ vary for each layer. For space considerations, only 3 convolutions are drawn here. In practice, there are 18 convolutional layers.}
\label{fig:synthesis}
\end{figure}

\begin{figure}
\begin{center}
\includegraphics[width=0.3\linewidth]{chart/AdaIN.jpg}
\end{center}
\caption{Each latent code is projected using a learned linear layer into a per-channel bias and offset which operate on the conv layer activations.}
\label{fig:adain}
\end{figure}

\fi

StyleGAN is a very recent architecture introduced by Karras et al. \cite{karras2018style}. StyleGAN borrows elements from the fast style transfer architecture of \cite{huang2017arbitrary}. We will review the core elements of the StyleGAN generator.

The generator consists of two parts:
\begin{enumerate}
\item Mapping Network: An $8$ layer fully connected mapping network, mapping between random multivariate normal vectors and global latent codes. In this paper, we will denote the codes sampled from the normal distribution $s$. The mapping network will be denoted $M()$. The latent code at the output of mapping network $M()$, will be denoted $z$ i.e. $z = M(s)$.    
\item Synthesis network: A DCGAN-like generator architecture with several crucial modifications: 1) The input to the lower convolutional layer is a learned fixed code, which we will denote $c_0$. 2)  The style code is fed into each convolutional layer. At each layer, the latent code $z$ is first projected by a linear layer into a set of per-channel factors and offsets. The factors and biases are used to multiply the output of each channel of the convolutional layer activations. This operation is called Adaptive Instance Norm (AdaIN), previously used in style transfer \cite{huang2017arbitrary} and in MUNIT \cite{huang2018multimodal}. In this paper, we denote the synthesis network as $G()$. The synthesis network takes the latent code $z$ as input  and synthesizes an image $\tilde{I}$ i.e. $\tilde{I} = G(z) = G(M(s))$.   
\end{enumerate}

StyleGAN is trained using adversarial training and progressive growing. We use the exact generator trained by the authors. For further details of StyleGAN's adversarial training procedure, please consult the original paper and the official code release.  

\subsection{Latent Optimization}
\label{sec:method:latent}

The task of latent optimization sets to invert the generator $G(z)$ given an image. More formally:

\begin{equation}
    \label{eq:invert}
    z^{*} = \arg\min_{z}\|\phi(G(z)) - \phi(I)\|_1
\end{equation}

Where $\phi()$ is a perceptual feature vector describing the contents of the image. 

Let us make precise the definition of compactness, sufficiency and invertibility made in the introduction. 

\begin{enumerate}
    \item Compactness: For a compact generative model, for every valid latent code $z$, the generated $G(z)$ should lie within the target image distribution $\cal T$.
    \item Sufficiency: For a sufficient generative model, for every image $I$ which comes from the target image distribution $\cal T$, there exists a valid latent code $z$, such that $G(z)=I$.
    \item Invertibility: For an invertible generative model, for every image $I$ from the target distribution  $\cal T$, we will be able to find a solution to Eq.~\ref{eq:invert} within a reasonable amount of computational effort.
\end{enumerate}

We note that a generator that is invertible, is necessarily sufficient, but the opposite is not true. 

A practical algorithm that we use to invert a generative model, first initializes the latent code with zeros and then directly optimizes Eq.~\ref{eq:invert} over the latent code until convergence. We use a learning rate of $0.001$ and $1000$ iterations.

\subsection{Inverting Style Generators}
\label{subsec:method:inv_style}

In this work, we show that style generators are invertible. There are several latent codes which we can invert in the StyleGAN architecture: i) The latent code  $s$ which is the input to the mapping network $M()$. The advantages are that its distribution should be simple and approximately Gaussian. The disadvantage is the large number of fully-connected layers that must be inverted. ii) The latent code $z$, which is the output of the mapping network $z=M(s)$ and input to the generator $G()$. The advantage is that it does not require inverting the mapping network $M()$. iii) Another possibility is learning a different latent code $z_l$ for every layer $l$. Although the same code is used for all layers during StyleGAN training, the attribute flipping regularization ensures that the generator does not assume the same latent code is applied to all layers.

In our experiments, we found that the extra degrees of freedom afforded by learning per-layer latent codes $z_l$ provide the best performance of all methods.

\subsection{Generative Models as Image Priors}
\label{subsec:method:prior}

In the previous sections, we described how to obtain high-quality generative models as well as effective techniques for inverting them. In this section, we describe how generative models can be used as strong image priors.

For generators that are compact and sufficient, we can make the following observation: if an image can be modelled by the generator then it comes from the target distribution, if it cannot be modelled by the generator then it does not. Although, this prior is binary in nature, and cannot distinguish images of different likelihoods, it is already very powerful. In practice, generators are not exactly compact or sufficient, and are certainly not perfectly invertible. Instead, we use the reconstruction error as the prior cost, as shown in Eq.~\ref{eq:prior} (where $\phi()$ is a VGG perceptual loss). We use the same implementation as in the public code release of \cite{nam_eccv}.    
\begin{equation}
    \label{eq:prior}
    L_{prior} = \min_{z}\|\phi(G(z)) - \phi(I)\|_1
\end{equation}

Smaller values of the cost function, imply an image that comes from the distribution $\cal T$, whereas large values imply out-of-distribution images. This cost therefore acts as an image prior.

Models that are not compact will have small reconstruction costs even for images that are out-of-distribution. Models that are not sufficient or invertible will have large reconstruction errors even for images that are in-distribution. To act as a good prior, a generative model therefore needs to be compact, sufficient and invertible. In the experimental section we show that style generators satisfy the conditions for being good generators. 

\subsection{Inverting Models for Disentanglement}
\label{sec:method:decomposable}

One of the attractive properties of generative image models is the linearization of semantic attributes in latent space. This can be observed by linear interpolations between two latent codes, which smoothly vary semantic attributes. Such behavior is not seen in pixel space, where linear interpolation does not result in smooth variation of the semantic attributes.

Let us assume that there exists a space in which semantic attributes are fully disentangled. In this section, the attributes that we consider are identity and pose. In the disentangled space, each latent code is therefore expressed as:

\begin{equation}
\label{eq:vector_id_p}
z = \begin{bmatrix}
           z_{id} \\
           z_p
         \end{bmatrix}
\end{equation}

Where $z_{id}$ is the identity part and $z_p$ is the pose part.

Even if we make the (reasonable) assumption that the learned latent space is indeed linear, we cannot assume that it is disentangled, but rather it could be any affine transformation of this space. Let us denote this affine transformation $W$. We decompose $W$ into its left and right columns:

 \begin{equation}
\label{eq:wid_wp}
W = \begin{bmatrix} W_{id} & W_p \end{bmatrix}
\end{equation}

The observed latent codes therefore become:

 \begin{equation}
\label{eq:}
\tilde{z} = Wz = W_{id} z_{id} + W_p z_p = \tilde{z_{id}} + \tilde{z_p}
\end{equation}

If we take the origin of pose to be at $z_p=0$, then for a sufficiently diverse set of latent code observations, the average $\langle \tilde{z} \rangle = \tilde{z_{id}}$. The per-frame pose code is given by $\tilde{z_p} = \tilde{z} - \langle \tilde{z} \rangle$. This pose vector can be used to transfer pose to the image of any other person by addition to the target person's identity latent code. In practice, we find that normalizing $\tilde{z_p}$ and multiplying by a constant factor yields the best performance. 

\section{Experiments}
\label{sec:exp}

We performed extensive experiments to verify our contributions:

\subsection{Reconstruction Experiments}
\label{subsec:rec}

We argued in Sec.~\ref{sec:intro} that image generators should be excellent image priors if only they were compact, sufficient and invertible. Previous GAN trained generative models have not been able to synthesize images from the evaluation set and often not even images used for training. This is due to the combination of mode-collapse and the difficulty of inverting the latent to image transformation. They have therefore not satisfied the invertibility and sufficiency properties, although they have typically satisfied compactness. 

In this section, we perform experiments illustrating that style-based generative models do indeed satisfy sufficiency and invertibility. 

The task that we compare in this section is reconstruction of out-of-domain test images. More precisely,  the task for generator $G(z)$ is to find latent code $\tilde{z}$ such that $G(\tilde{z)}$ is as similar as possible to test image $I$. The optimization task becomes:

\begin{equation}
    \label{eq:rec}
    L_{rec} = \|\phi(G(\tilde{z})) - \phi(I)\|_p
\end{equation}
Where $\phi()$ is a pre-trained feature representation and $p$ is the order of the metric. We optimize the latent code $z$ using VGG perceptual features as implemented in \cite{nam_eccv}. For evaluation, we follow the LPIPS protocol, to avoid training exactly on the testing measure. At evalution time, the feature representation $\phi()$ is taken from LPIPS \cite{zhang2018unreasonable}. The loss metric is $L_1$.

We compare our proposed method: latent optimization of the StyleGAN generator, against two baselines: Progresive Growing of GANs (PGGAN) \cite{karras2017progressive}, \cite{celeba} and the generator trained by the GitHub repository of Mescheder et al. \cite{Mescheder2018ICML}. The two baseline methods were both trained on the CelebA-HQ dataset \cite{karras2017progressive} while StyleGAN is trained on FlickrFaces-HQ (FFHQ). The baseline GANs are selected to represent the previous state-of-the-art face generators. 

\begin{figure}
\begin{center}
\includegraphics[width=0.24\linewidth]{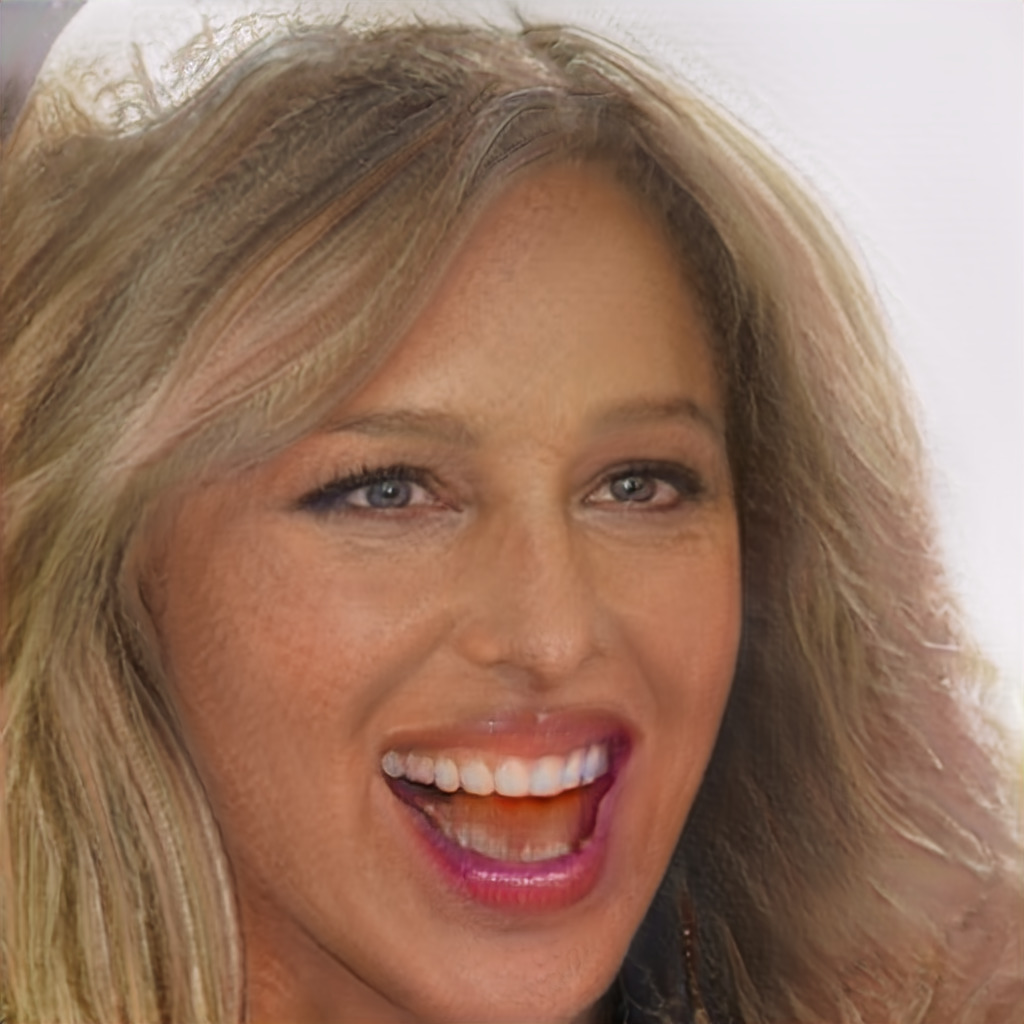}
\includegraphics[width=0.24\linewidth]{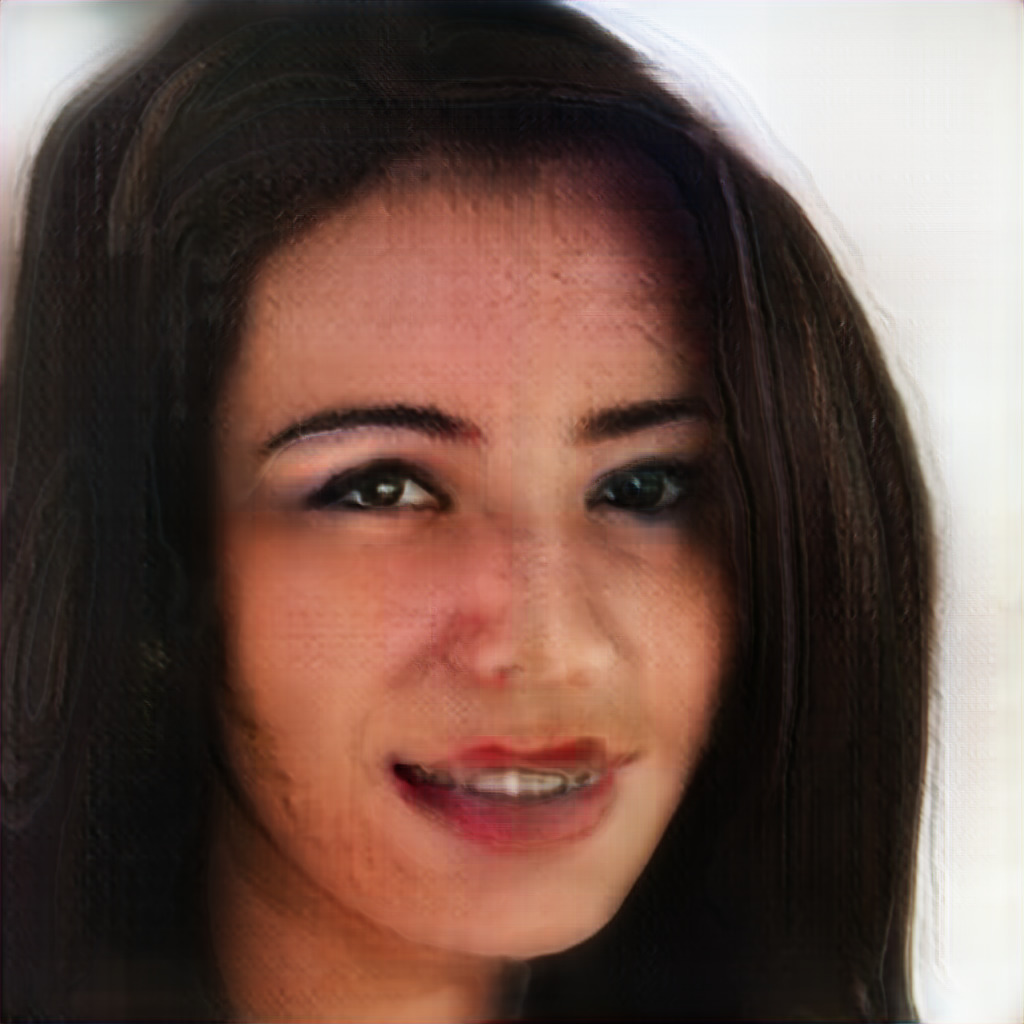}
\includegraphics[width=0.24\linewidth]{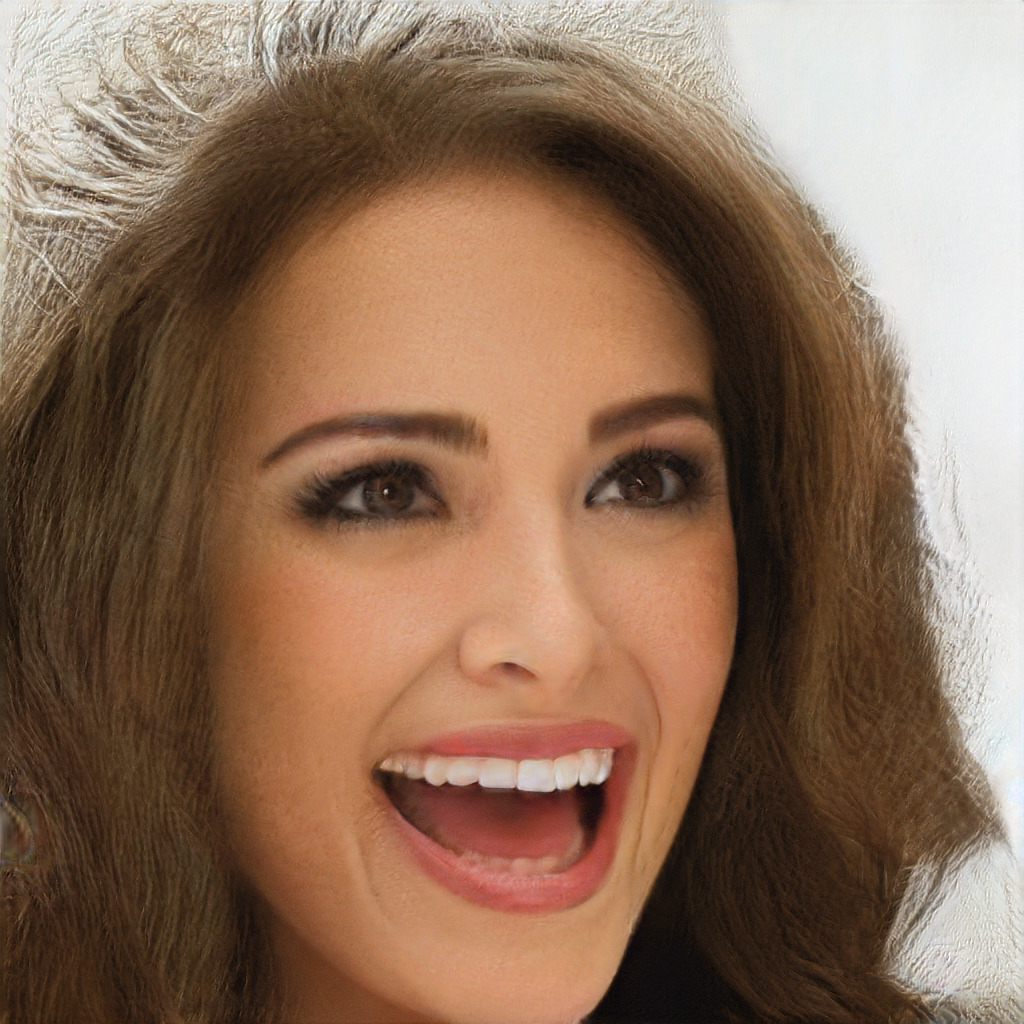}
\includegraphics[width=0.24\linewidth]{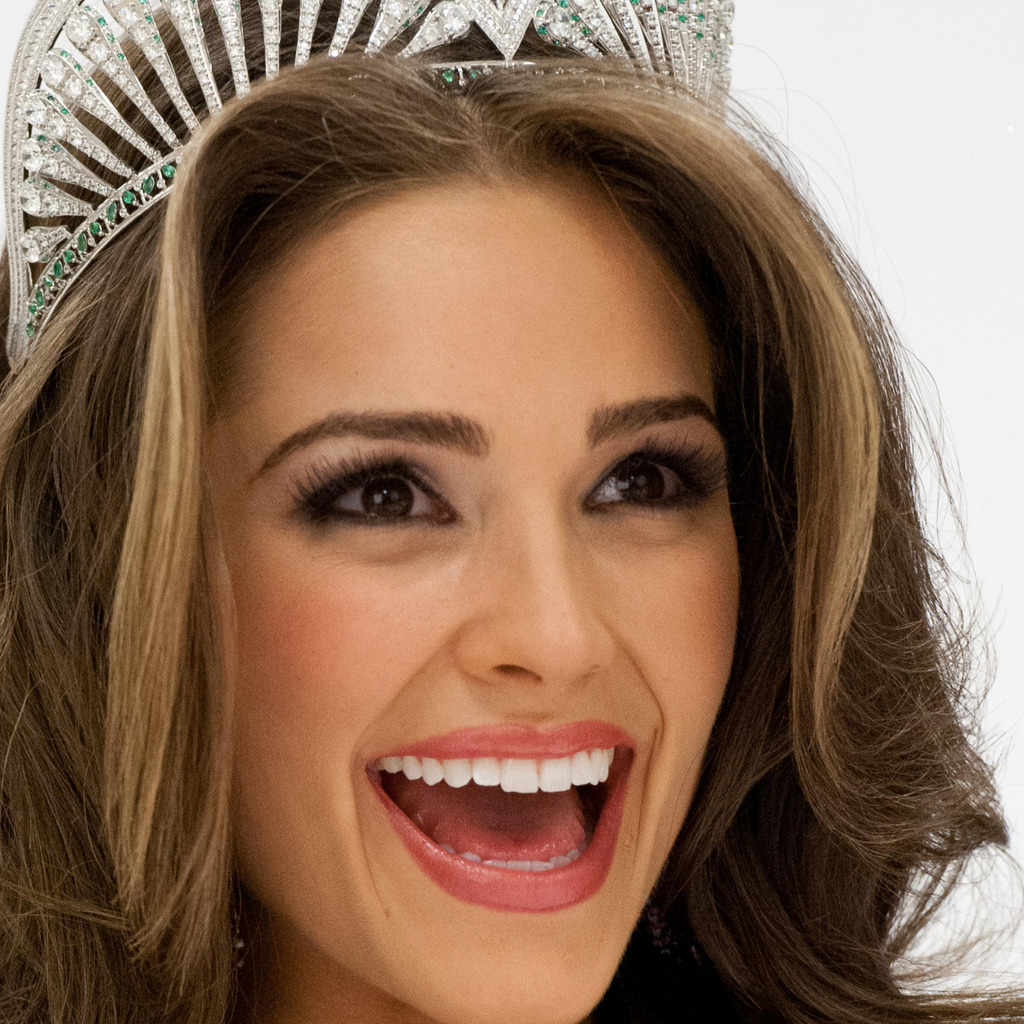}
\end{center}
\caption{A reconstruction of an image from CelebA-HQ. (right-to-left PGGAN, Mescheder et al. \cite{Mescheder2018ICML}, StyleGAN, Groundtruth). Inverting StyleGAN preserves both style and identity properties in contrast to other GAN-trained generators.}
\label{fig:reconstruction}
\vspace{-1em}
\end{figure}

We  present a qualitative comparison of the different methods in Fig.~\ref{fig:reconstruction}. Several important observations are apparent.  We can see that non-style generators are not-invertible that is, they perform poorly on in-distribution reconstructions (let alone out-of-distribution reconstruction). Style-generators perform well both in and out-of-distribution. 

\begin{table}[t]
  \centering
      
  \caption{Reconstruction Error for Different GANs (LPIPS)}
  \label{tab:rec_gans}

    \begin{tabular}{ccc}
    \toprule
    PGGAN & \cite{Mescheder2018ICML} & StyleGAN \\
    \midrule
   0.599 & 0.654 & \textbf{0.345}\\
	 \bottomrule
    \end{tabular}
\end{table}

We compare all methods on reconstructing the first 100 images from CelebA-HQ. In this case StyleGAN is trained out-of-distribution. A numerical comparison between the reconstruction errors of the different generators can be observed in Tab.~\ref{tab:rec_gans}. The numerical experiments confirm the observations we made in the qualitative experiments. 

\begin{figure}
\begin{center}
\includegraphics[width=0.45\linewidth]{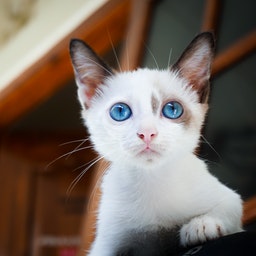}
\includegraphics[width=0.45\linewidth]{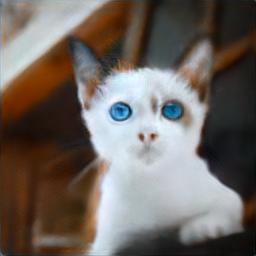}
\includegraphics[width=0.45\linewidth]{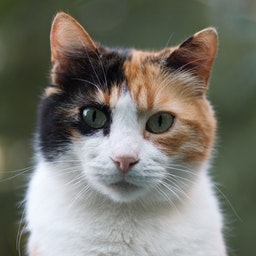}
\includegraphics[width=0.45\linewidth]{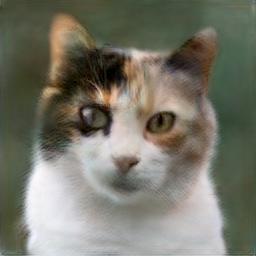}
\end{center}
\caption{Images of cats crawled from the web (left) and their reconstructions (right) using latent optimization of the cats StyleGAN model.}
\label{fig:reconstruction_other}
\vspace{-1em}
\end{figure}

To confirm that the invertibility is not simply limited to face generative models, we performed additional qualitative experiments on reconstruction of cat images. We used the weights of the StyleGAN generators trained on LSUN cats in \cite{karras2018style}. The qualitative reconstruction results are presented in Fig.~\ref{fig:reconstruction_other}. We can see that the perceptual quality of the reconstruction is very high, even on non-facial images.

As discussed in Sec.~\ref{subsec:method:inv_style}, there are alternative strategies for latent code inversion: i) Noise: latent optimization of the single-latent code $s$ prior to being passed through the mapping network $M()$, ii) Global: latent optimization of a post-mapping latent code $z$ that is replicated and inserted into each of the convolutional layers. iii) Per-Layer: latent optimization of a per-layer latent code $z_l$ . Note that the last two strategies do not use the mapping network at all (even though it was used for training the style-generator in the first place).

\begin{table}[t]
  \centering
      
  \caption{Reconstruction Error for Different Optimizations (LPIPS)}
  \label{tab:rec_latents}

    \begin{tabular}{ccc}
    \toprule
    Noise & Global & Per-Layer \\
    \midrule
  0.842 & 0.398 & \textbf{0.345} \\
   
	 \bottomrule
    \end{tabular}
\end{table}

In Tab.~\ref{tab:rec_latents}, we calculated the average reconstruction loss for the first 100 images in CelebA-HQ for the three methods. It confirms that per-layer latent optimization is better than the other methods. 

\subsection{Image Prior Experiments}
\label{subsec:prior}

In Sec.~\ref{subsec:rec}, we established that differently from previous generative models, style-generators are invertible and sufficient. In this section, we evaluate the effectiveness of style-generators as priors for image processing inverse-problems.

\begin{figure*}
\begin{center}
\includegraphics[width=0.16\linewidth]{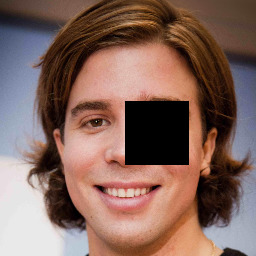}
\includegraphics[width=0.16\linewidth]{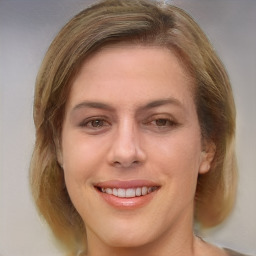}
\includegraphics[width=0.16\linewidth]{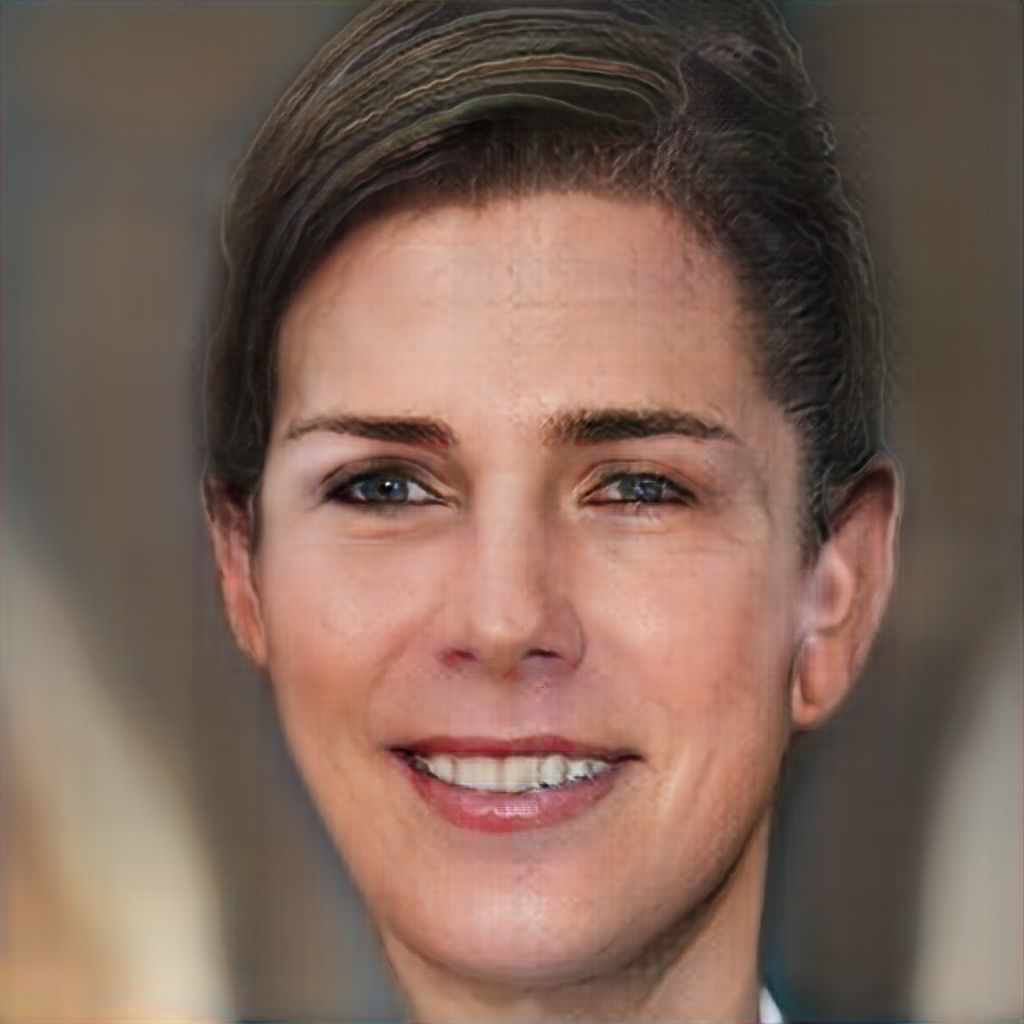}
\includegraphics[width=0.16\linewidth]{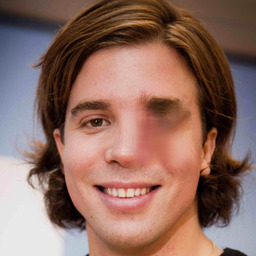}
\includegraphics[width=0.16\linewidth]{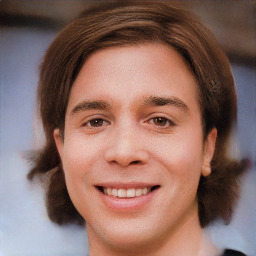}
\includegraphics[width=0.16\linewidth]{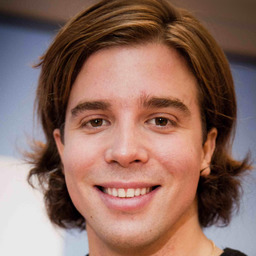}
\includegraphics[width=0.16\linewidth]{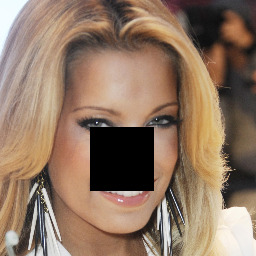}
\includegraphics[width=0.16\linewidth]{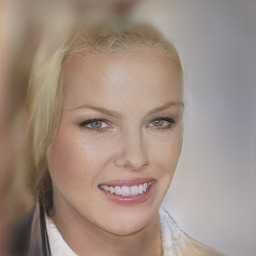}
\includegraphics[width=0.16\linewidth]{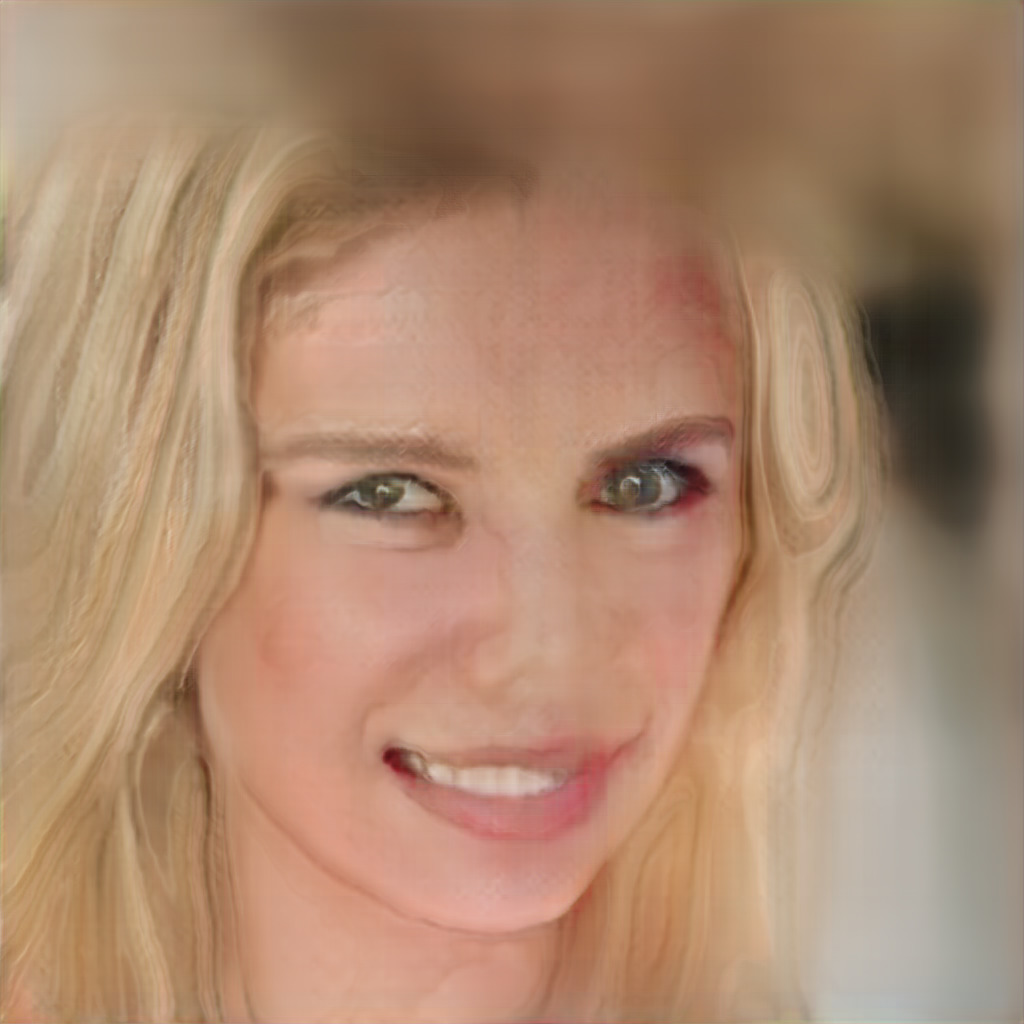}
\includegraphics[width=0.16\linewidth]{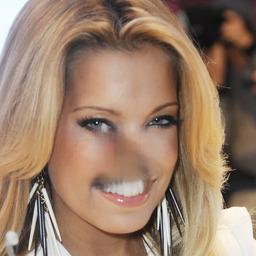}
\includegraphics[width=0.16\linewidth]{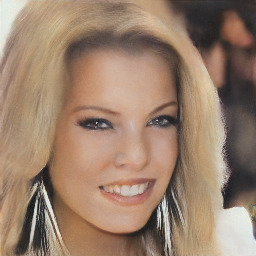}
\includegraphics[width=0.16\linewidth]{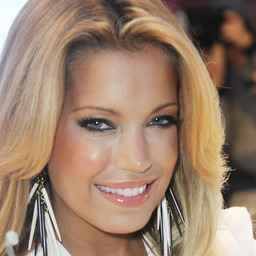}
\end{center}
\caption{Image inpainting results on two example images (From left: Corrupted, PGGAN, Mescheder et al. [22], DIP, Ours, GT). PGGAN and \cite{Mescheder2018ICML} cannot maintain the identity, DIP does not have anatomical knowledge, whereas our method can achieve both.}
\label{fig:inpainting}
\vspace{-1em}
\end{figure*}

\begin{figure*}
\begin{center}
\includegraphics[width=0.16\linewidth]{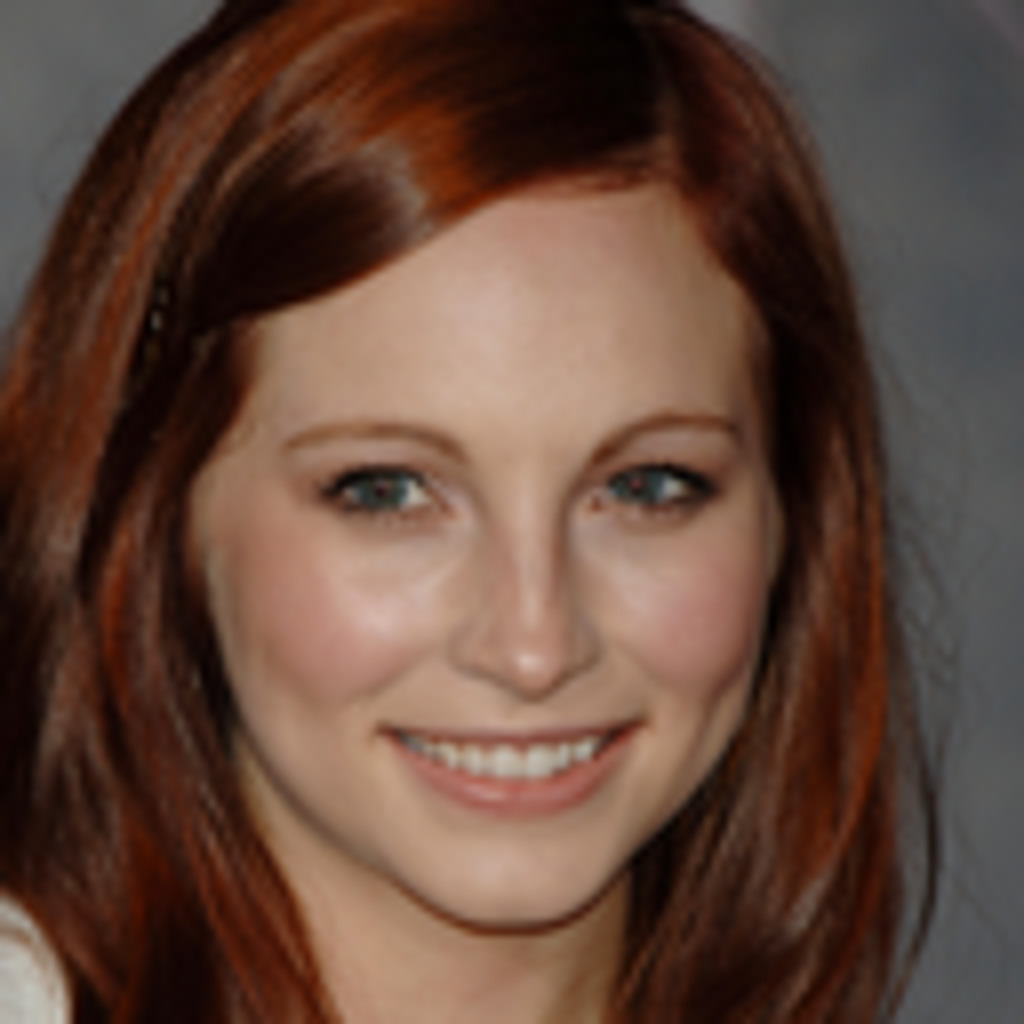}
\includegraphics[width=0.16\linewidth]{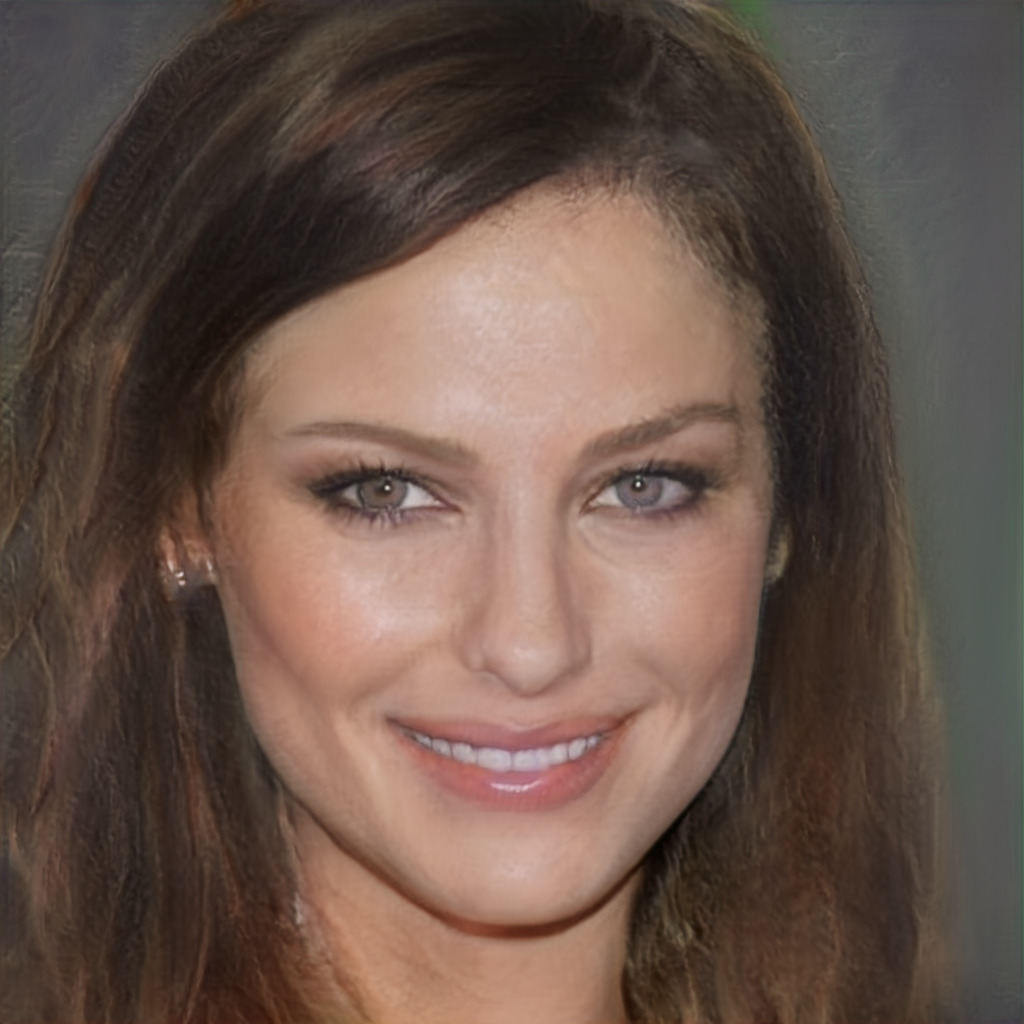}
\includegraphics[width=0.16\linewidth]{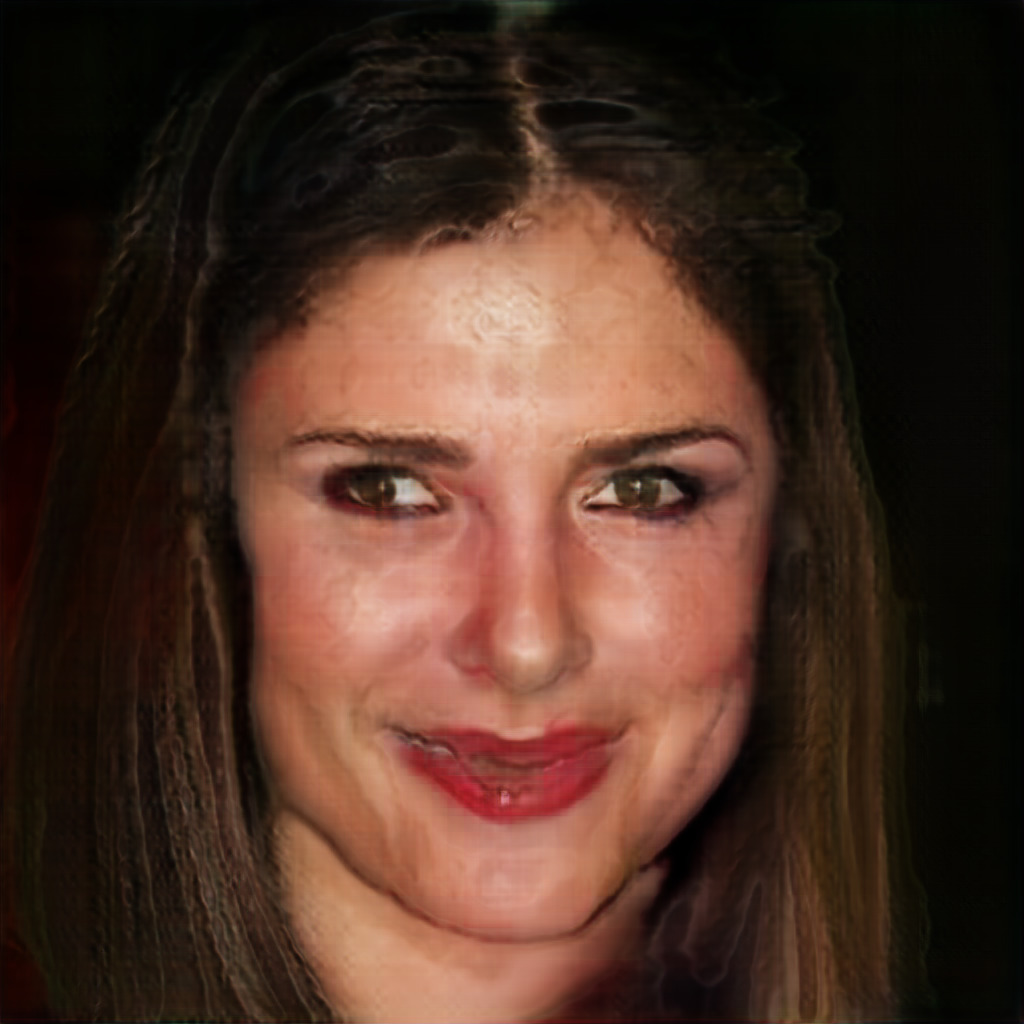}
\includegraphics[width=0.16\linewidth]{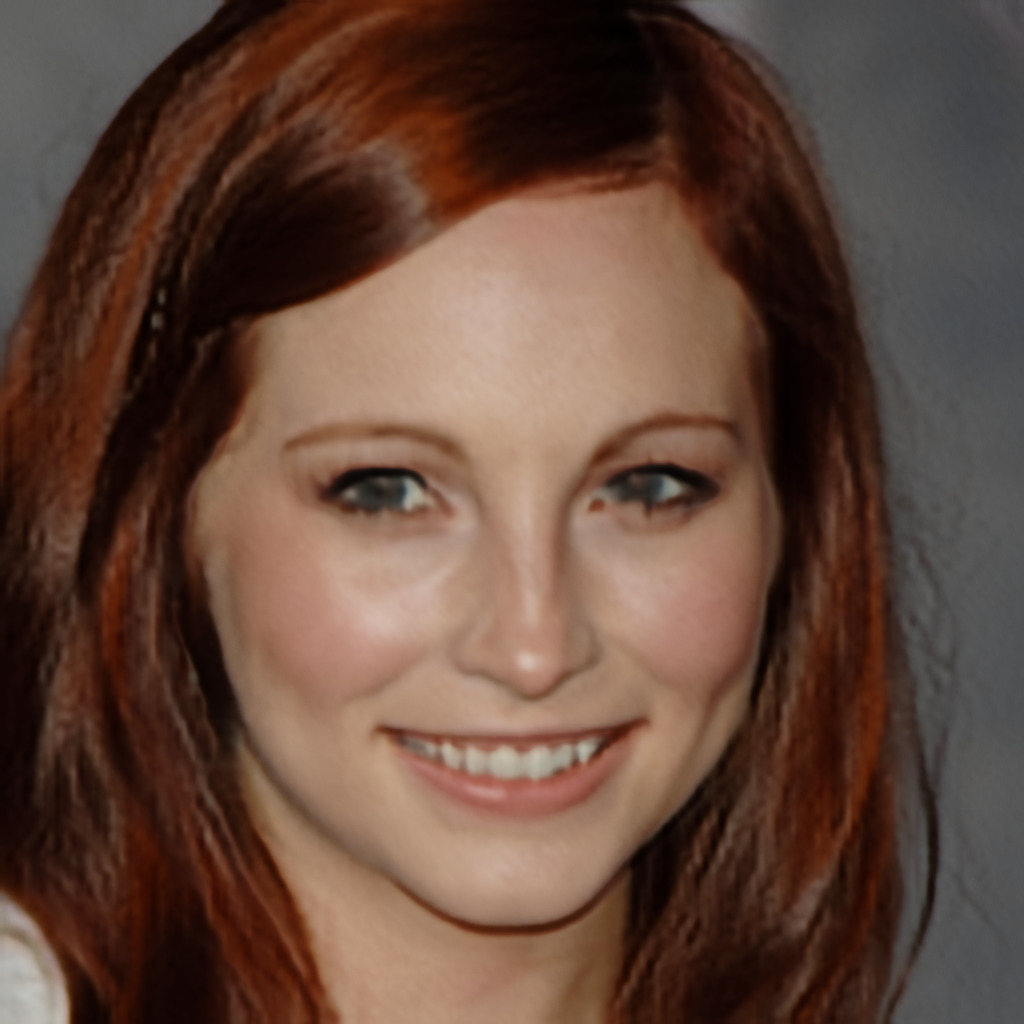}
\includegraphics[width=0.16\linewidth]{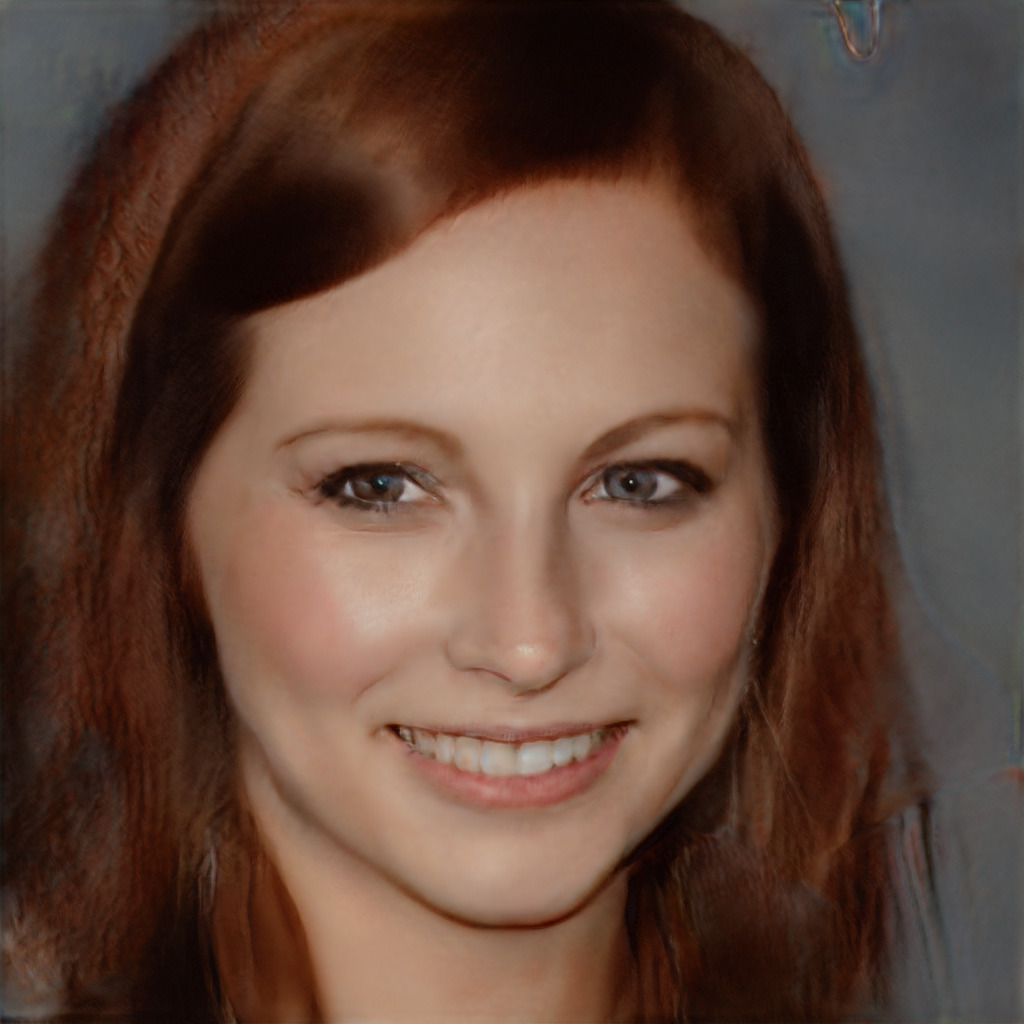}
\includegraphics[width=0.16\linewidth]{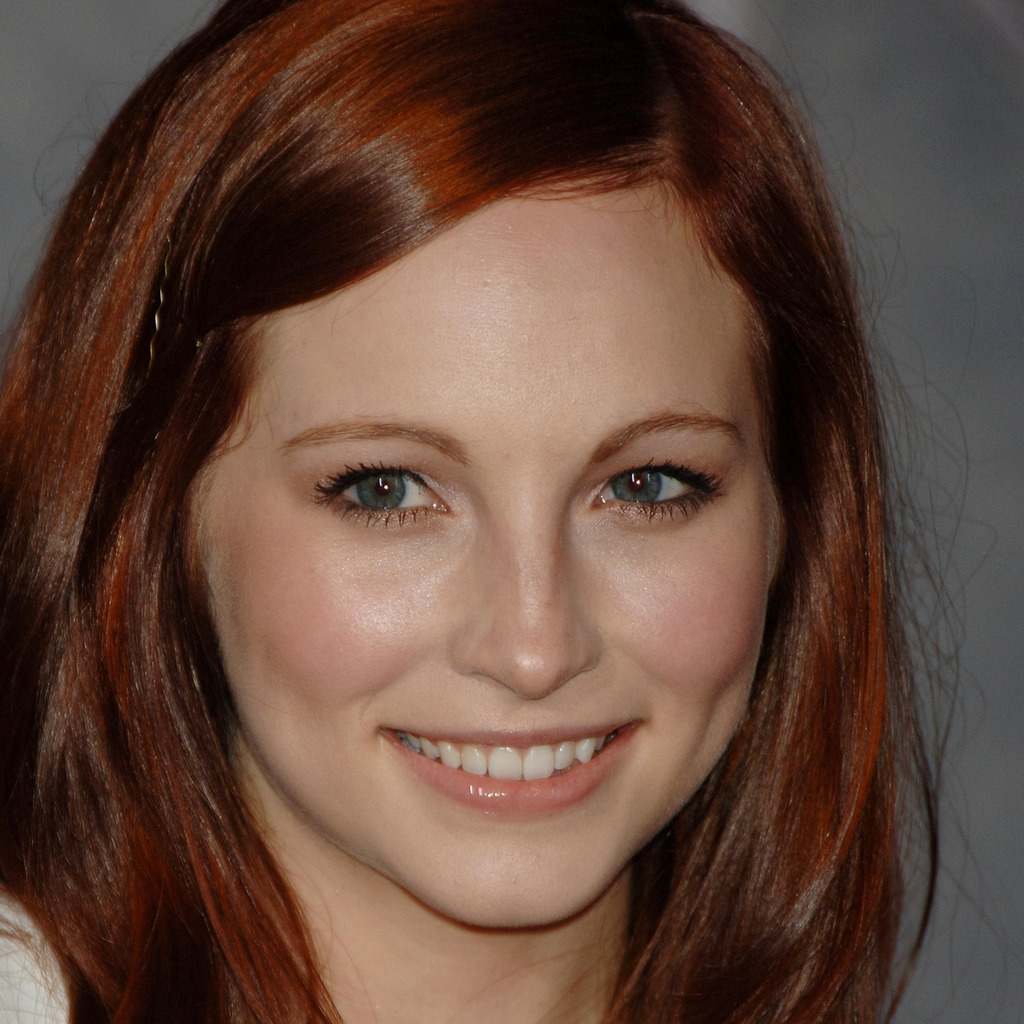}

\includegraphics[width=0.16\linewidth]{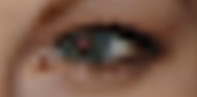}
\includegraphics[width=0.16\linewidth]{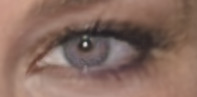}
\includegraphics[width=0.16\linewidth]{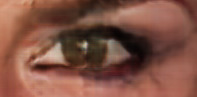}
\includegraphics[width=0.16\linewidth]{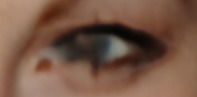}
\includegraphics[width=0.16\linewidth]{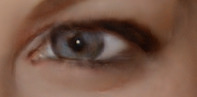}
\includegraphics[width=0.16\linewidth]{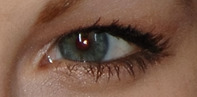}

\includegraphics[width=0.16\linewidth]{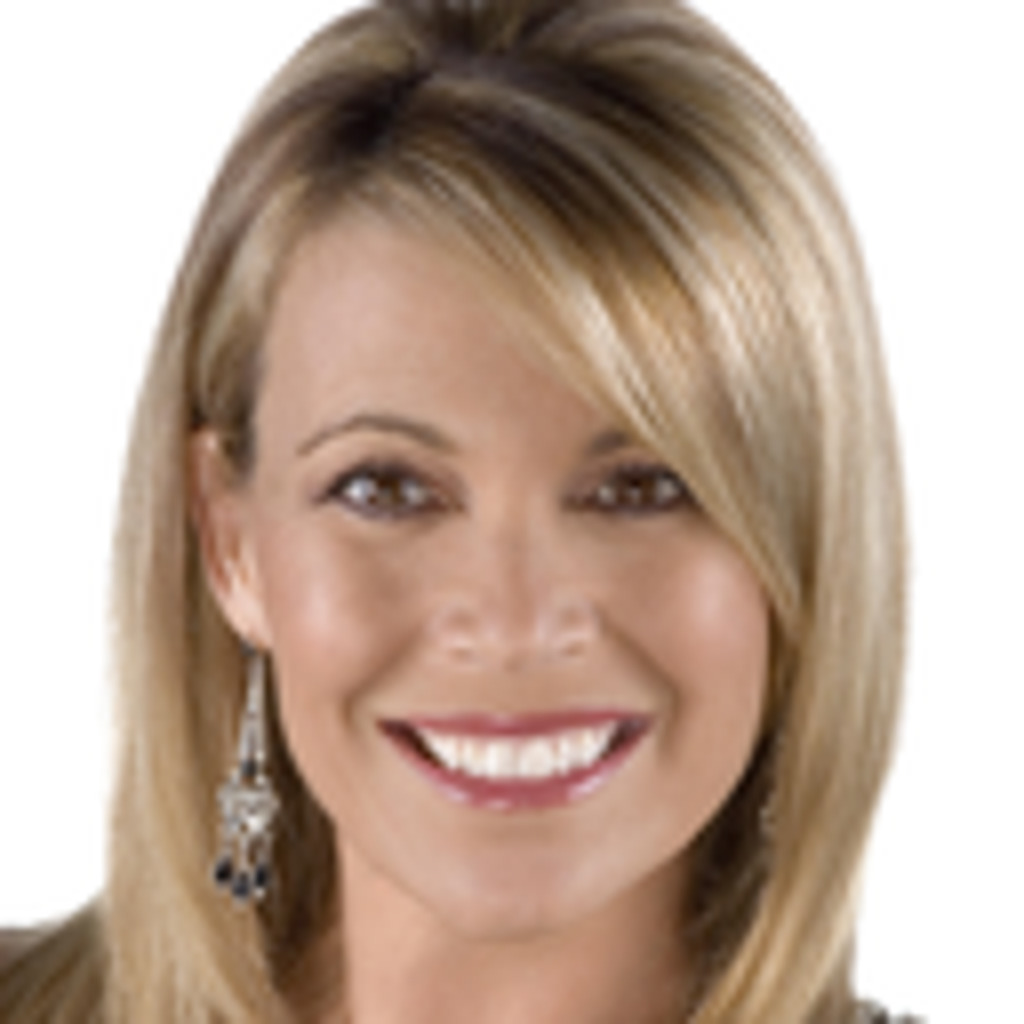}
\includegraphics[width=0.16\linewidth]{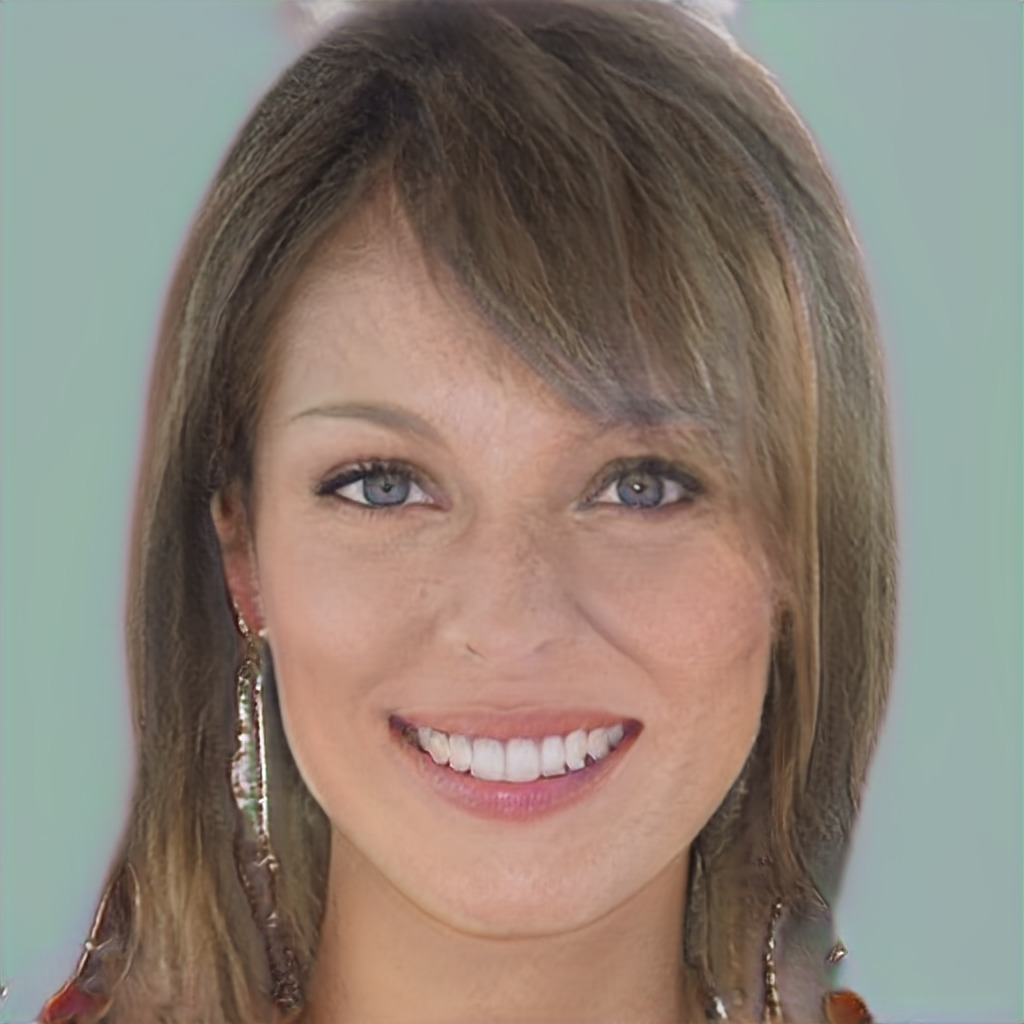}
\includegraphics[width=0.16\linewidth]{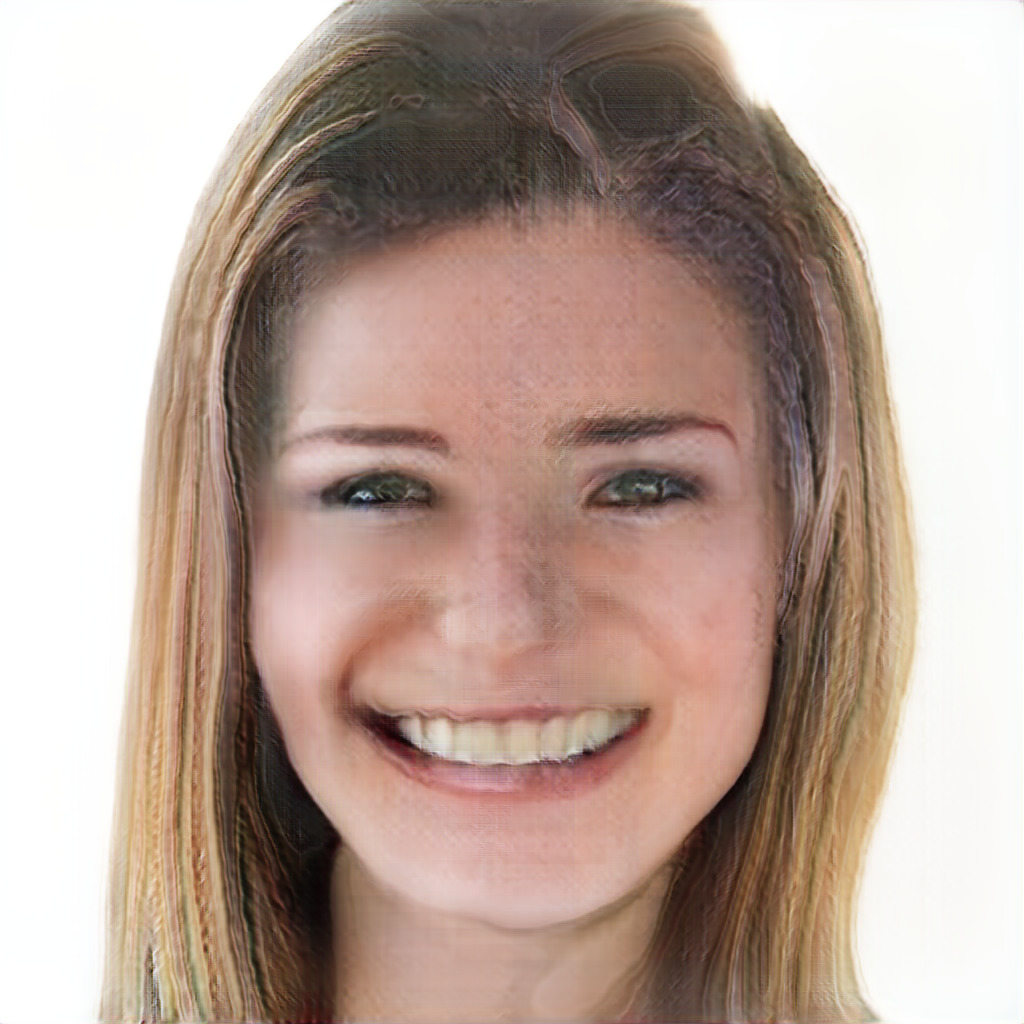}
\includegraphics[width=0.16\linewidth]{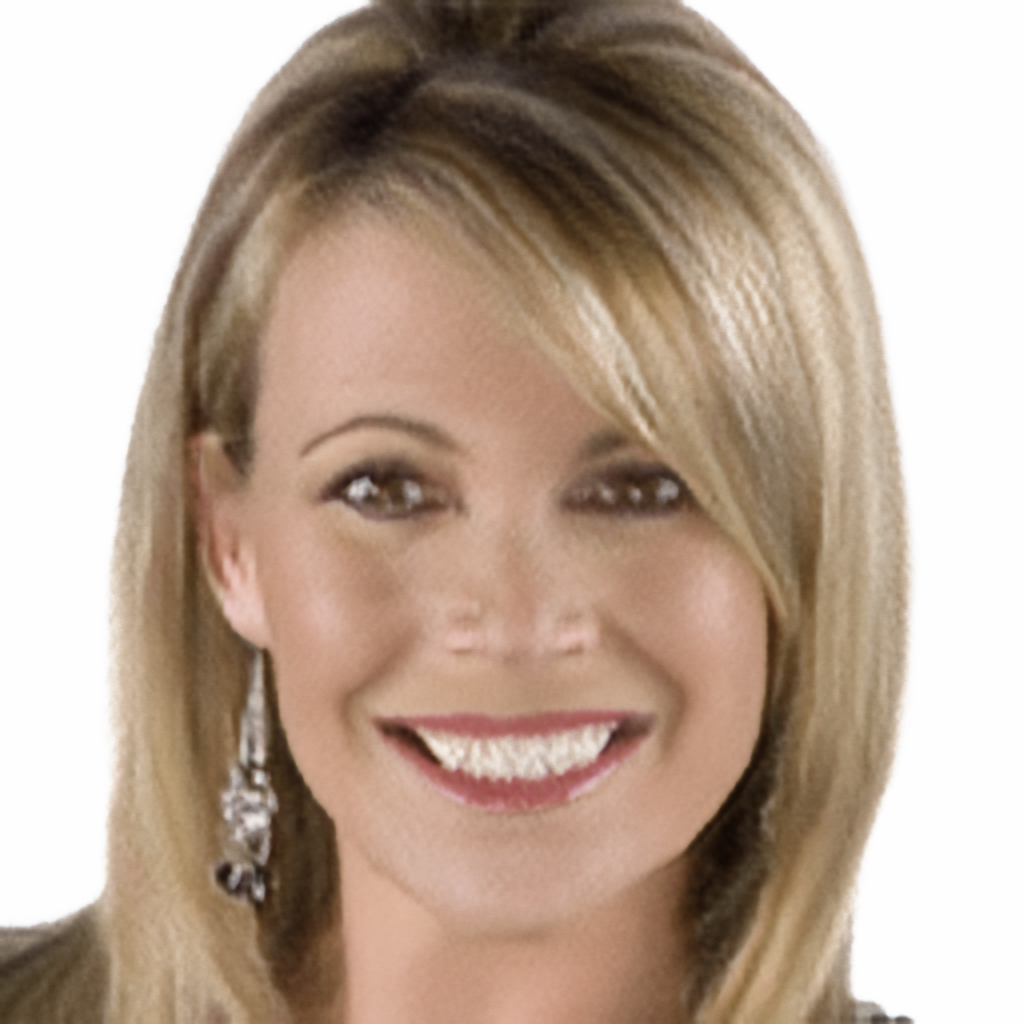}
\includegraphics[width=0.16\linewidth]{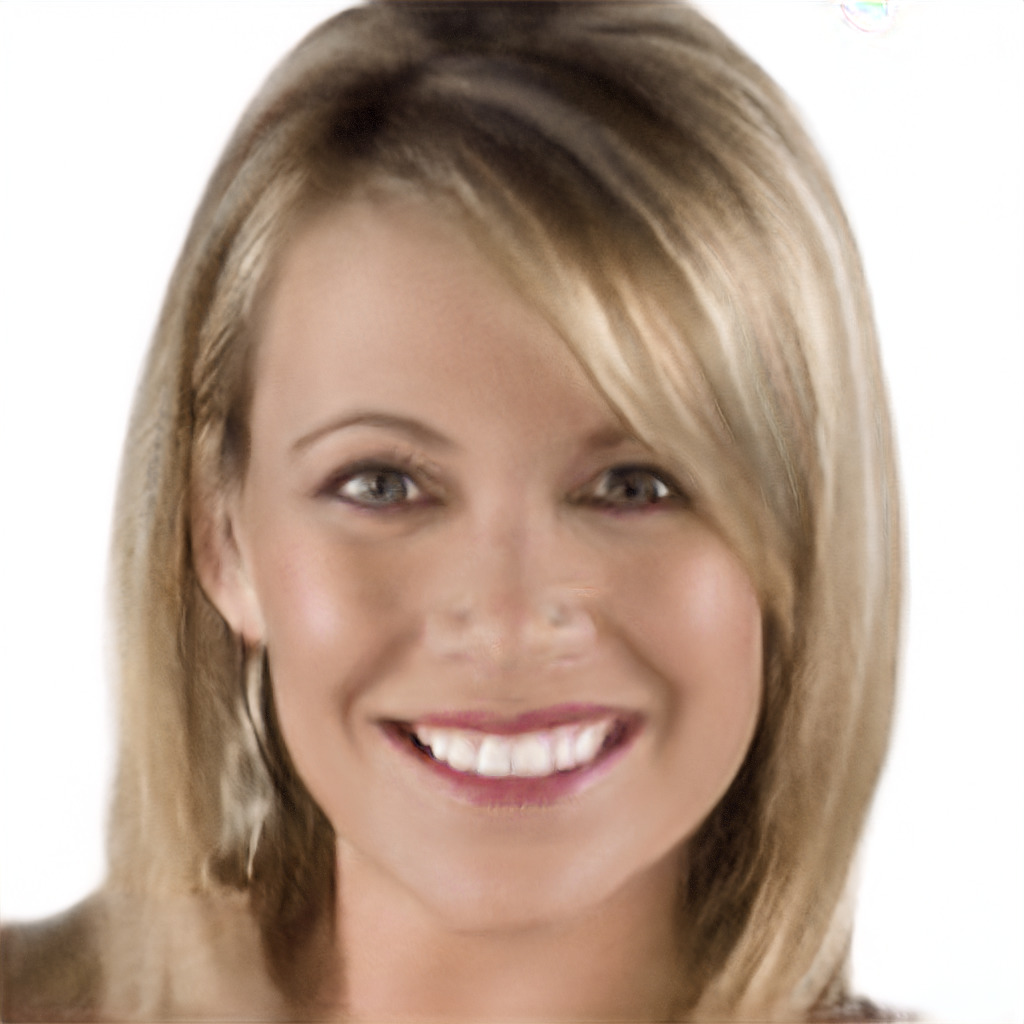}
\includegraphics[width=0.16\linewidth]{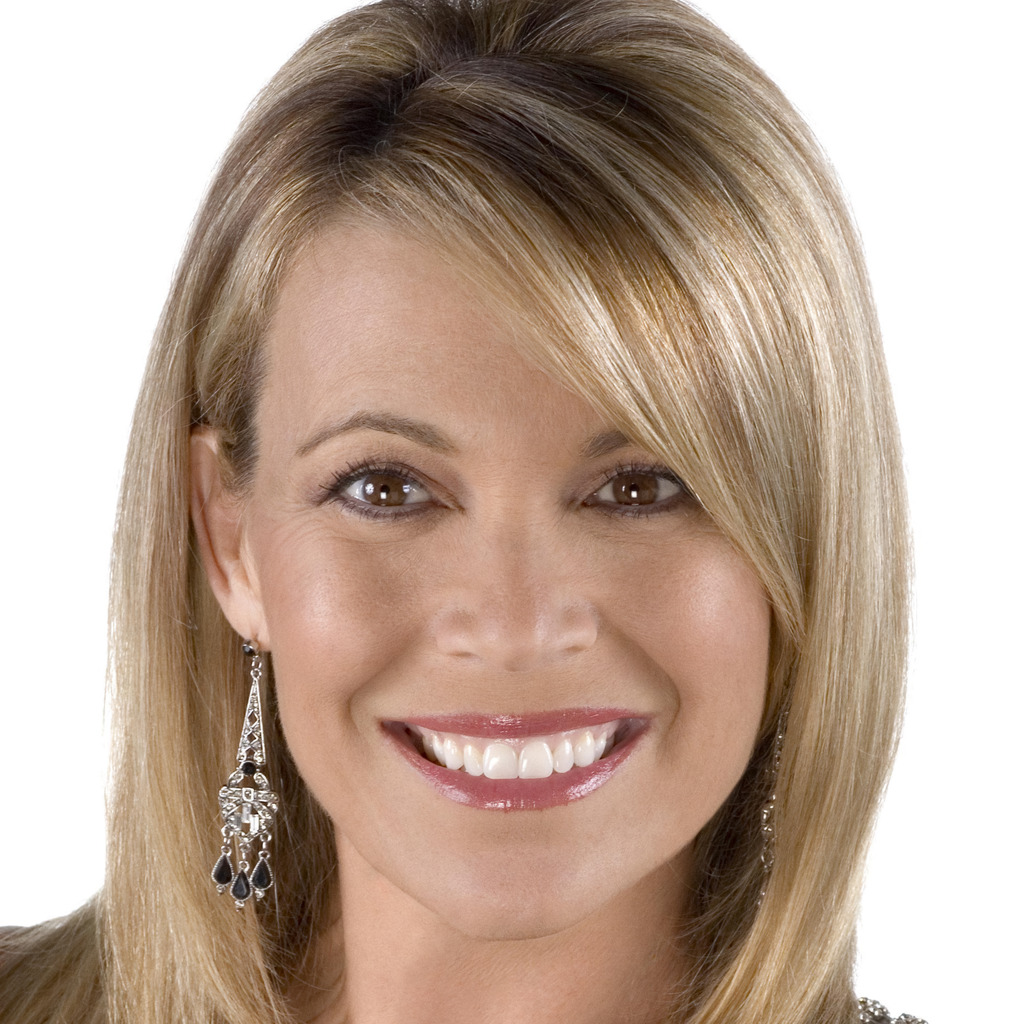}

\includegraphics[width=0.16\linewidth]{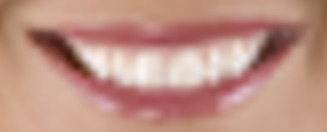}
\includegraphics[width=0.16\linewidth]{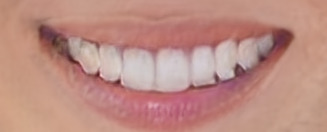}
\includegraphics[width=0.16\linewidth]{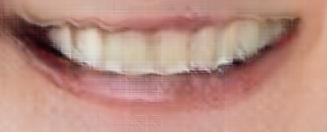}
\includegraphics[width=0.16\linewidth]{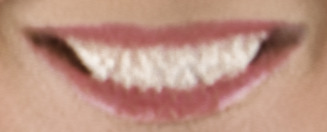}
\includegraphics[width=0.16\linewidth]{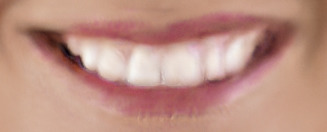}
\includegraphics[width=0.16\linewidth]{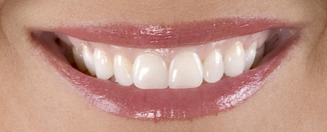}

\end{center}
\caption{Super-resolution results on two example images (From left: Bicubic, PGGAN, Mescheder et al. \cite{Mescheder2018ICML}, DIP, Ours, GT). PGGAN cannot maintain the identity, DIP cannot produce fine-features, our method performs well. We also provide crops showing the enhancement that our method provides on eyes and teeth.}
\label{fig:super_resolution}
\vspace{-1em}
\end{figure*}

\textbf{Image Inpainting:} Inpainting missing parts of an image is a challenging and long standing task. In this evaluation, we will concentrate on unsupervised inpainting that relies on having a strong image prior. 

In this task, the input image is multiplied by a known binary mask $m$, so that only some of the true pixels are observed. As the missing pixels can take arbitrary values, an image prior is needed to infer the values of the missing pixels. 

Mathematically the task is written in Eq.~\ref{eq:inpaint_task}, where $m$ is the binary mask:
\begin{equation}
    \label{eq:inpaint_task}
    L_{inpaint} = \|\phi(m \odot G(\tilde{z})) - \phi(m \odot I)\|_1
\end{equation}

The supposition is that given a perfect invertible image generator, the optimal latent code $\tilde{z}$ will be found such that the generated image $G(\tilde{z})=I$, i.e. that the generated image will generate the full image, filling in the correct details.

We use a binary mask $m$ with a square of size $64 \times 64$ missing pixels inserted at a random location (for each of the $100$ images a random mask was generated and used for all compared methods). We use DIP, PGGAN, Mescheder et al. and StyleGAN trained on FFHQ (which is out-of-distribution) to solve this task. The images were downscaled to $256 \times 256$ resolution, as the official DIP implementation could not handle larger resolutions. 

We compared the effectiveness of style-generative priors against other generative priors:

\begin{itemize}
\item Deep Image Prior: This method trains a convolutional neural network from scratch to reconstruct the target image. In \cite{ulyanov2018deep} the authors found that the method typically is able to synthesize images that appear be realistic. This approach does not use any domain knowledge. Although a powerful tool, we test whether not using any domain knowledge weakens the quality of the prior.

\item GAN-trained Generative Image Priors: We use state-of-the-art trained generators as image priors. As during training the generators had access to a large number of images, such models have domain knowledge. We established in the above section that such generators are often not reversible, we therefore test their suitability as image priors. Specifically we used PGGAN \cite{karras2017progressive} and Mescheder et al. \cite{Mescheder2018ICML} trained on CelebA-HQ as baselines.

\end{itemize}

We performed experiments on the first $100$ CelebA-HQ images. Examples of inpainting performance on two masked images can be observed in Fig.~\ref{fig:inpainting}. DIP is not able to complete the missing part of the face, as it requires knowledge of the anatomy of human faces, which is not contained in the image. The GAN baselines have not been able to fit to the image, owing to their lack of invertibility. Our method has quite faithfully reconstructed the true clean image.  

\begin{table}[t]
  \centering
      
  \caption{Inpainting Error for Different Priors (LPIPS)}
  \label{tab:inpaint_priors}

    \begin{tabular}{cccc}
    \toprule
    DIP & PGGAN & \cite{Mescheder2018ICML} & StyleGAN \\
    \midrule
   0.368 & 0.134 & 0.232 & \textbf{0.104} \\
   
	 \bottomrule
    \end{tabular}
\end{table}

% Aviv: should we note that our lpips in inpainting is much better than in regular reconstruction perhaps because we do not calculate it on the background?

Numerical results can be observed in Tab.~\ref{tab:inpaint_priors}. We measure the image similarity using the perceptual loss LPIPS, rather than using $L_1$ or $L_2$ losses which do not correspond to perceptual image dissimilarities. We can observe that DIP performs poorly on this task. Our method performs much better than all methods owing to its invertibility.

\textbf{Image Super-Resolution:} Another task commonly used to evaluate the performance of image priors is image super-resolution. In this task, the input image is downsampled by a factor of $8$, from $1024 \times 1024$ to $128 \times 128$. We use the different image priors to infer the missing details lost by the downsampling operation. Mathematically the task is written in Eq.~\ref{eq:sr_task}, where $D_8()$ is an operation that downsamples an image by a factor of $8$:
\begin{equation}
    \label{eq:sr_task}
    L_{SR} = \|\phi(D_8(G(\tilde{z}))) - \phi(D_8(I))\|_1
\end{equation}

As above, we compared our approach against DIP, PGGAN, \cite{Mescheder2018ICML} and a standard bicubic interpolation. DIP used $64$ rather than $128$ channels in the code release to ensure that it fits on GPU memory at $1024 \times 1024$ resolution. 

Results on two images are in Fig.~\ref{fig:super_resolution}. We observe that the baseline GANs have typically failed to fit to the image distribution and have lost the identity of the low-resolution face. This is striking as they were specifically trained on these images. DIP hallucinated unrealistic details, whereas StyleGAN was able to hallucinate reasonable details.

A quantitative comparison of LPIPS scores evaluated on the first $100$ images of CelebAHQ is presented in Tab.~\ref{tab:sr_priors}. In this case we observe that our method significantly outperformed other GANs and has also performed better than DIP.

\begin{table}[t]
  \centering
      
  \caption{Super-Resolution Error for Different Priors (LPIPS)}
  \label{tab:sr_priors}

    \begin{tabular}{cccc}
    \toprule
    DIP & PGGAN & \cite{Mescheder2018ICML} & StyleGAN \\
    \midrule
    0.419 & 0.573 & 0.626 & \textbf{0.375} \\
   
	 \bottomrule
    \end{tabular}
\end{table}

% \textbf{Results on other datasets:} We present further qualitative inpainting and super-resolution experiments, on the Cars and Cats datasets. The images were taken from the internet, and were not used for training. We can see that the good performance of our method is not limited to faces, but is further validated in non-face datasets.

\subsection{Application: Image Re-animation}
\label{subsec:reanimation}

\begin{figure*}
\begin{center}
\includegraphics[width=0.18\linewidth]{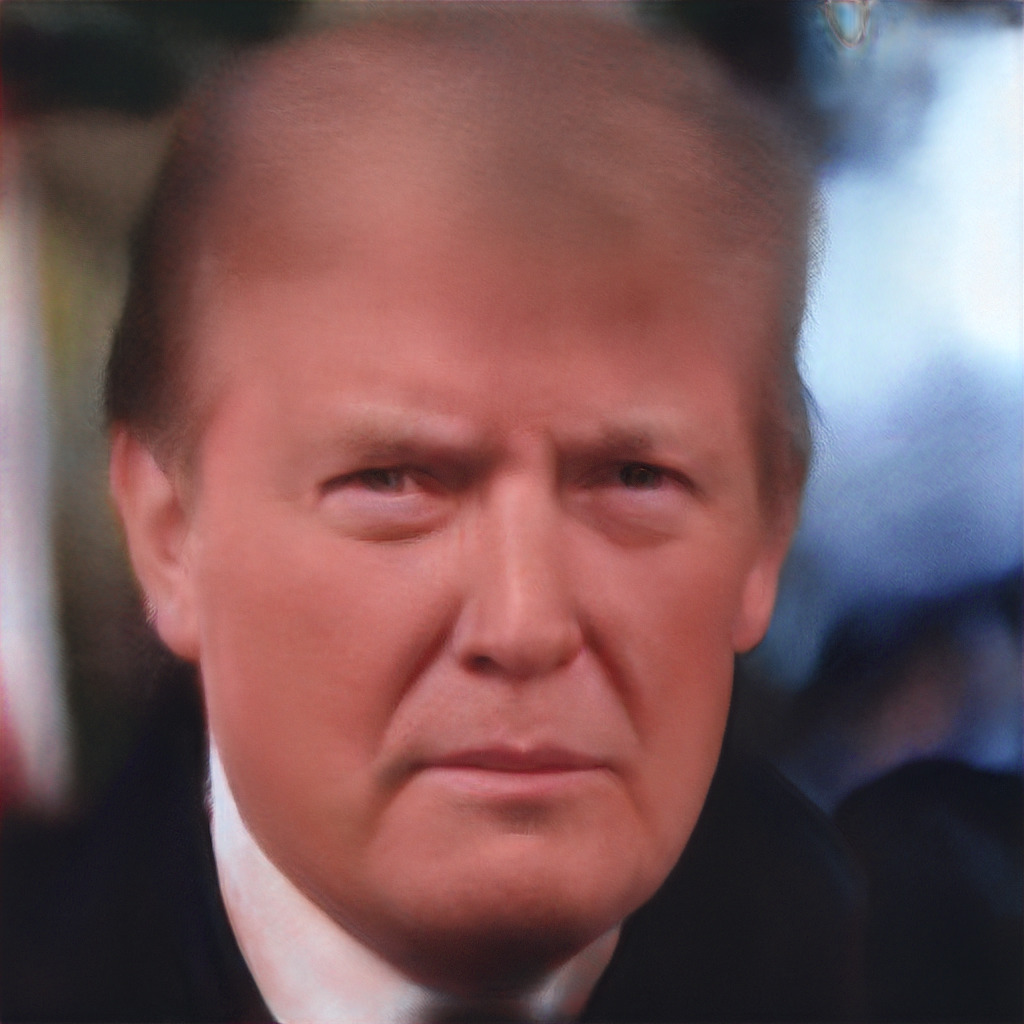}
\includegraphics[width=0.18\linewidth]{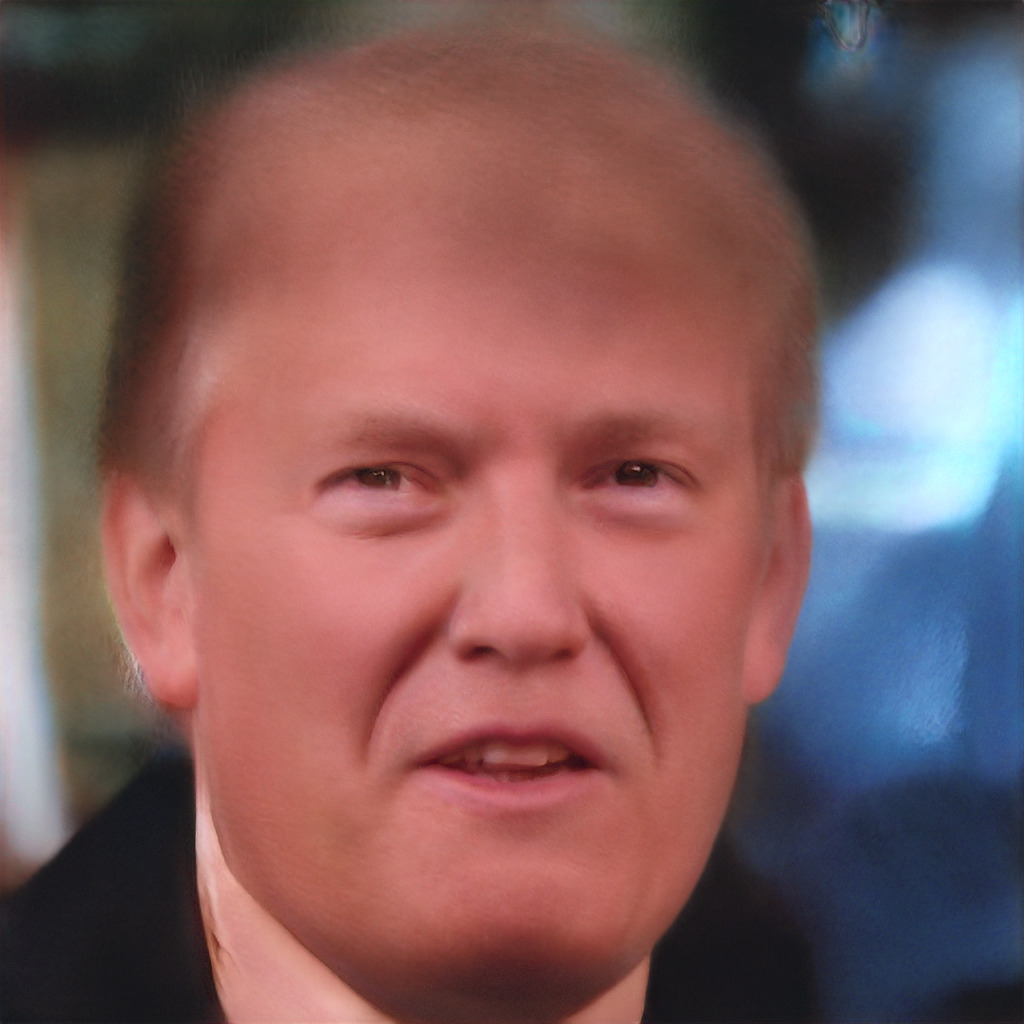}
\includegraphics[width=0.18\linewidth]{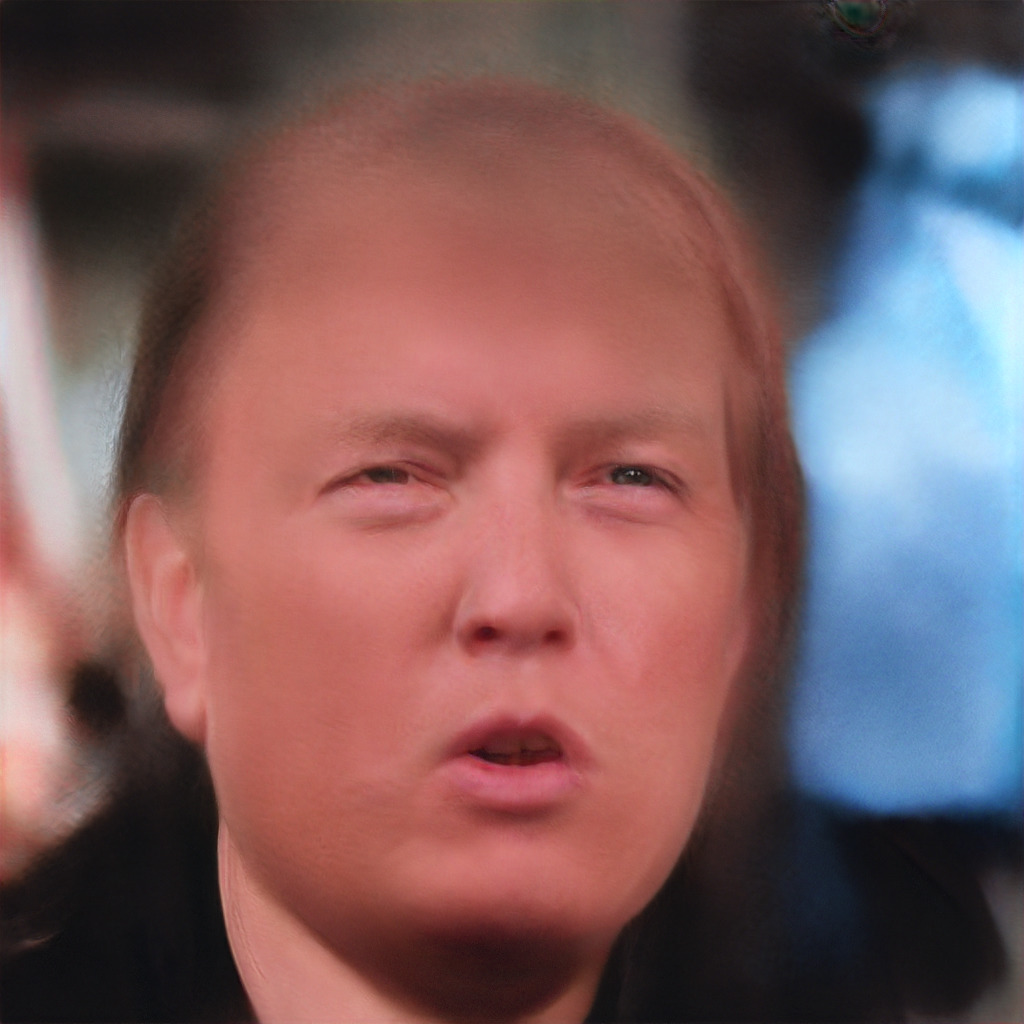}
\includegraphics[width=0.18\linewidth]{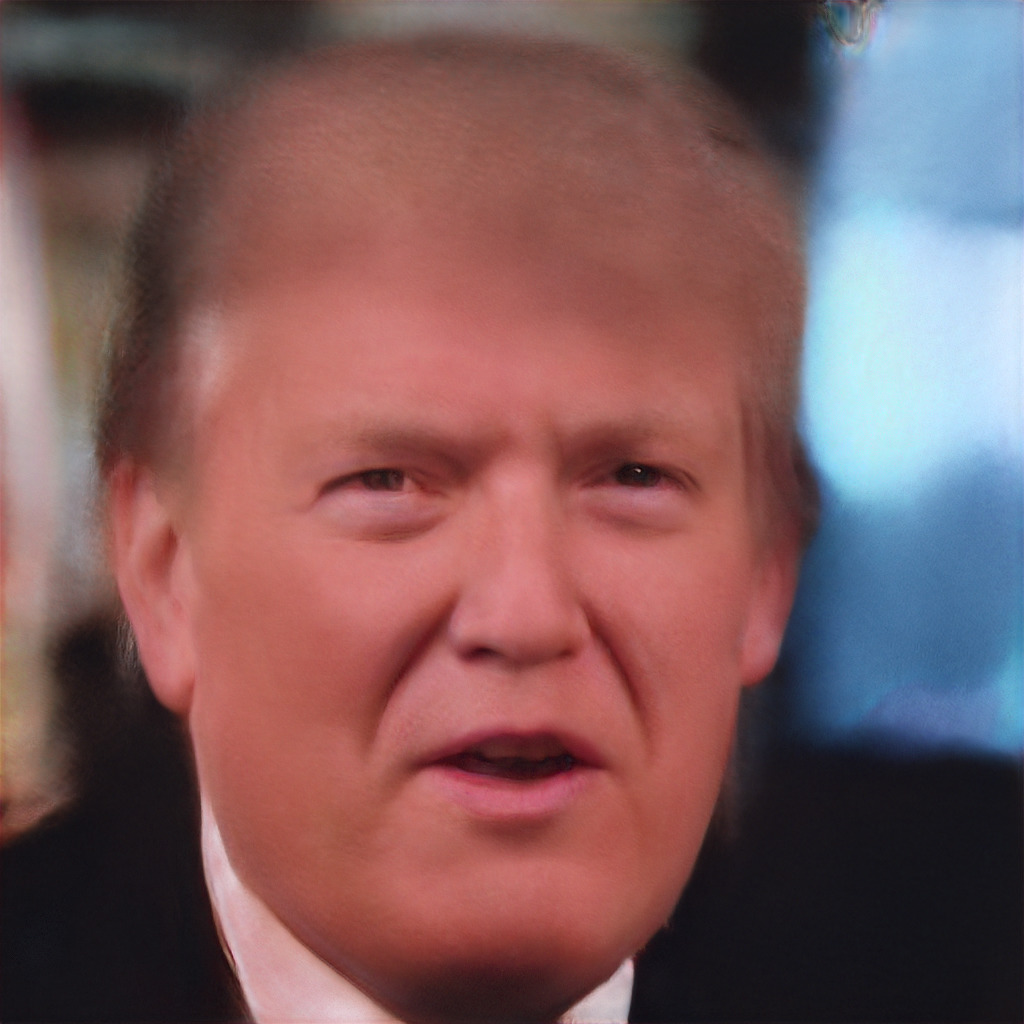}
\includegraphics[width=0.18\linewidth]{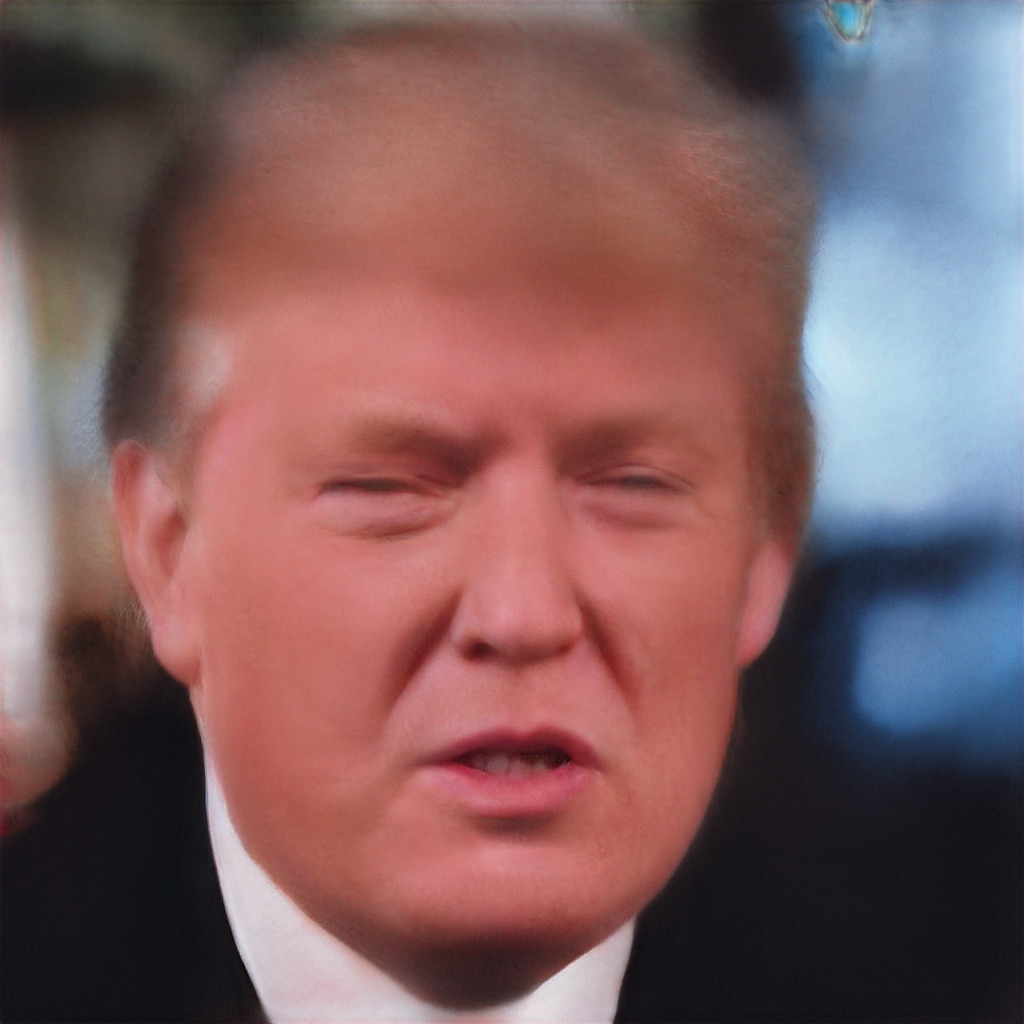}

\includegraphics[width=0.18\linewidth]{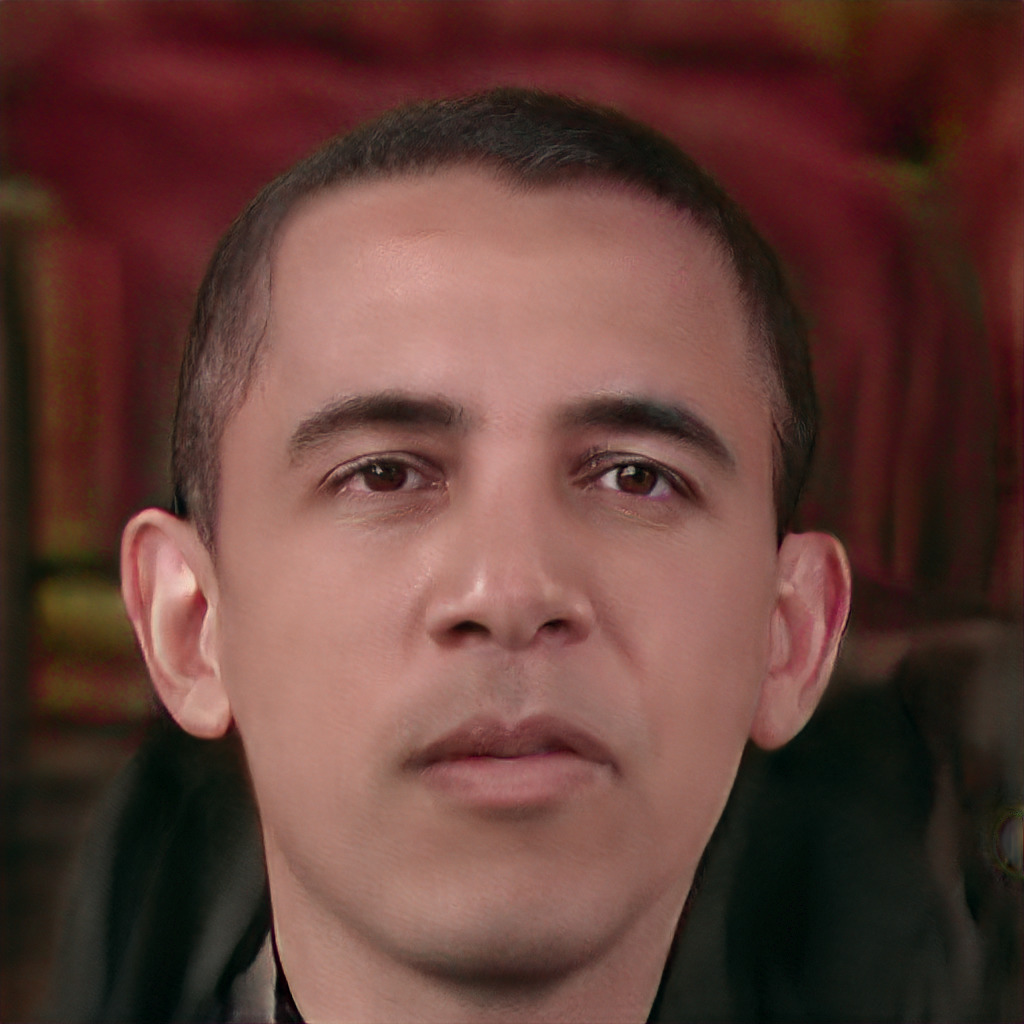}
\includegraphics[width=0.18\linewidth]{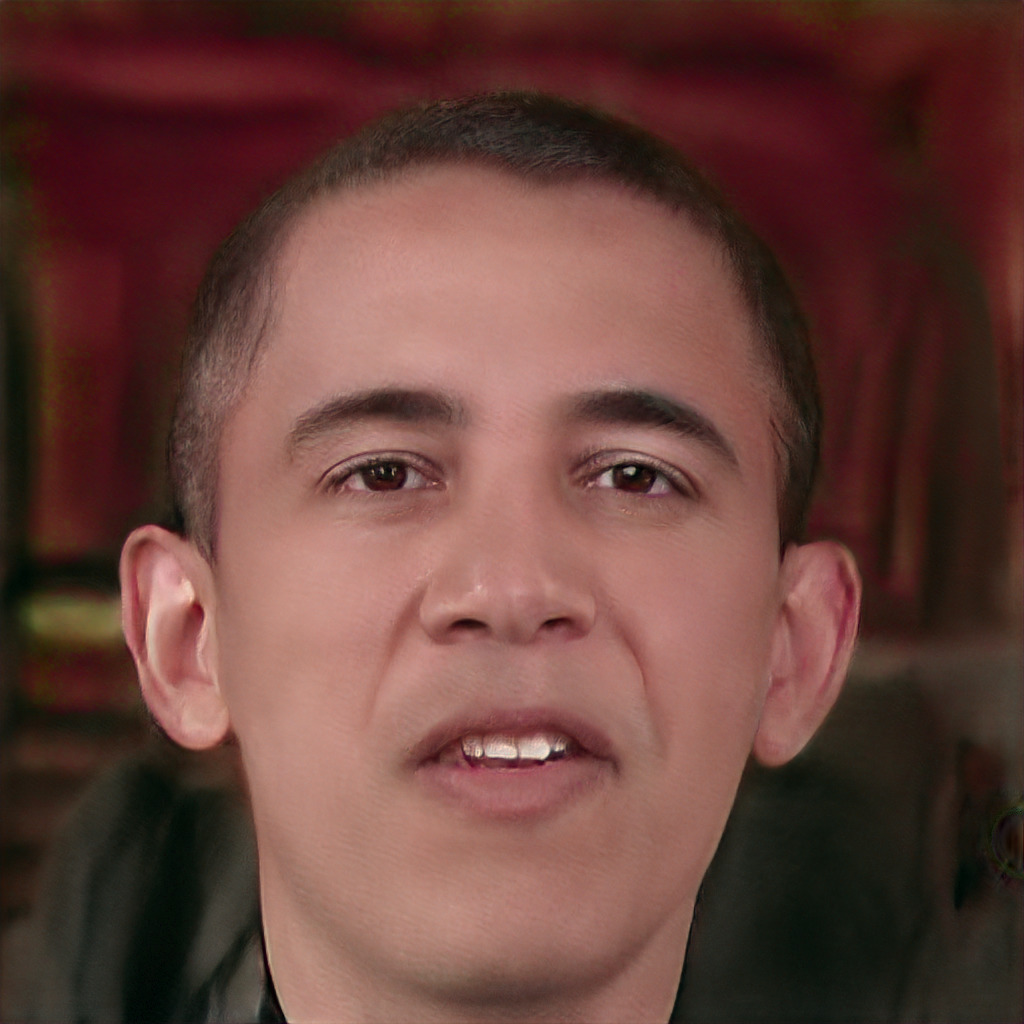}
\includegraphics[width=0.18\linewidth]{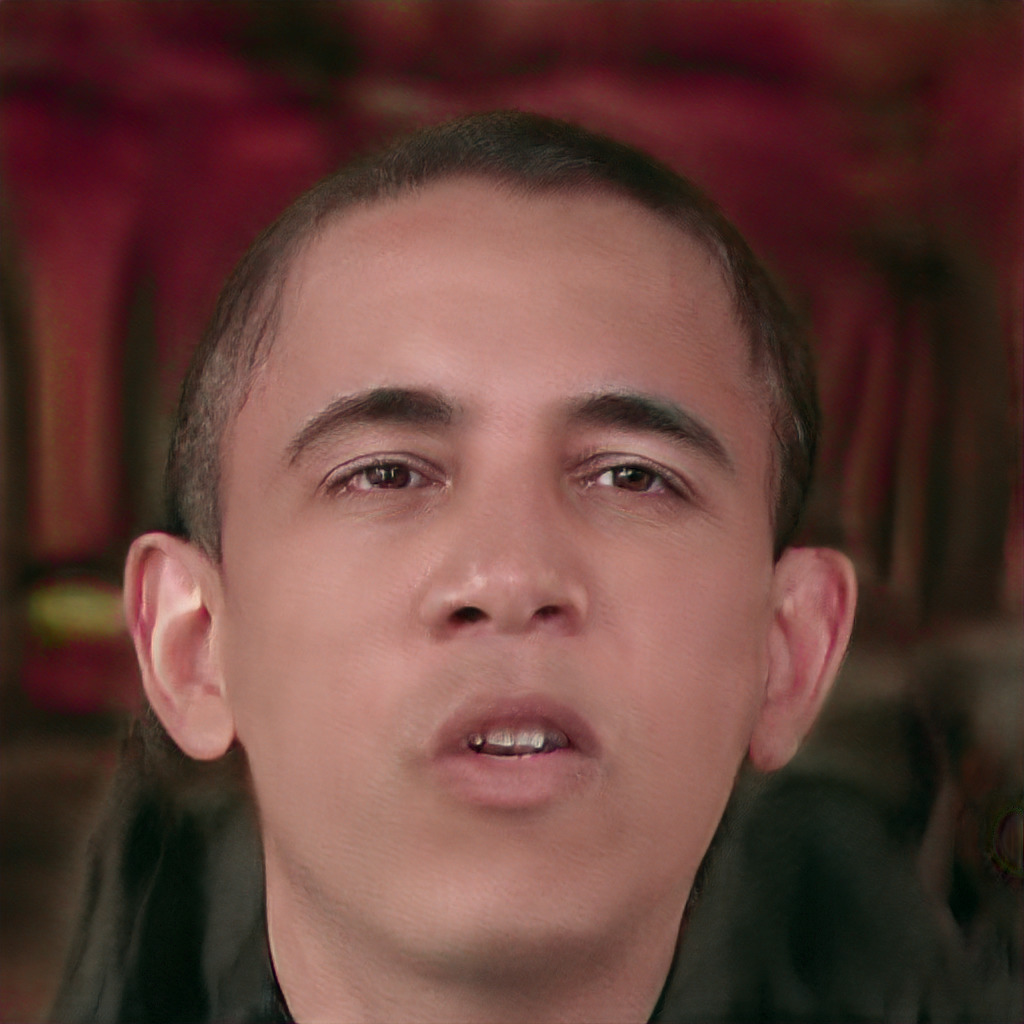}
\includegraphics[width=0.18\linewidth]{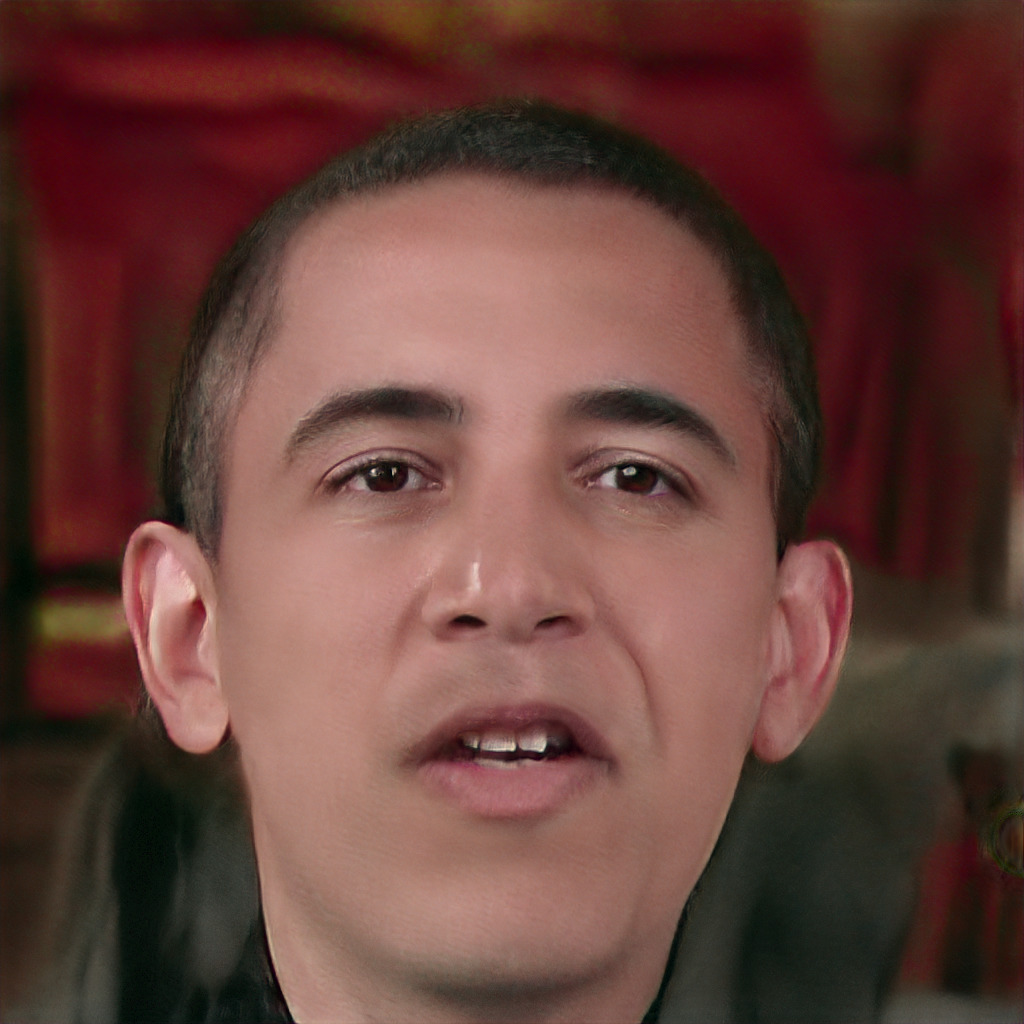}
\includegraphics[width=0.18\linewidth]{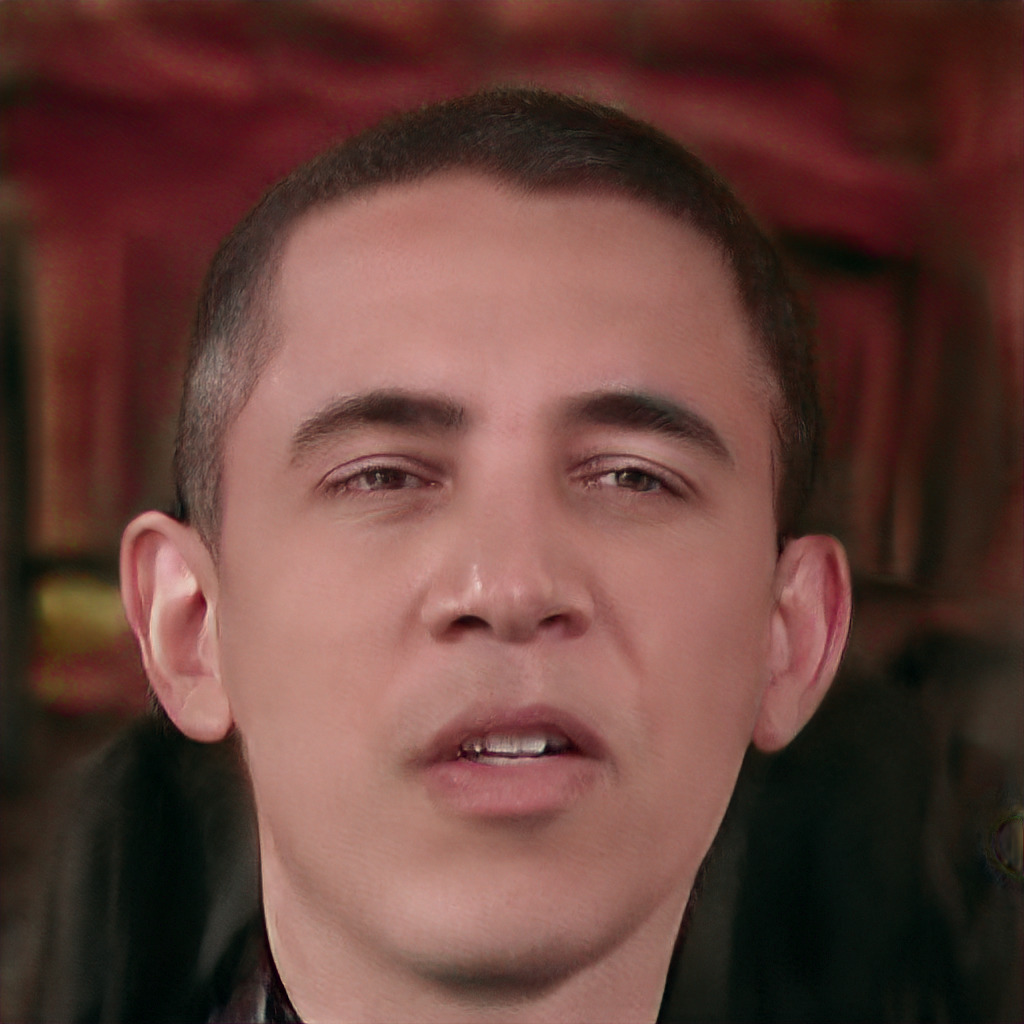}

\end{center}
\caption{We present reconstructions of $5$ frames from a video of Trump. For each frame, we compute the pose code and added it to Obama's identity code. The pose transferred results are present at the bottom row.}
\label{fig:animation}
\vspace{-1em}
\end{figure*}

Animating still images has been an object of fascination for centuries. Recently, success was achieved on the task of face reanimation, mostly by supervised methods which make use of fiducial points \cite{elor2017bringingPortraits} or action units \cite{pumarola2018ganimation}. In this section, we show that the invertibility of generators can be successfully utilized for unsupervised animation of images.

To illustrate the performance of our method, we transfer facial animation from a video of Trump to a still image of Obama. We first perform latent optimization on each image from a video of a speech by Trump. Each source image is encoded as a single latent code $z^s_i$. We then average all latent codes, to obtain an approximate source identity latent code $z^s_{id}$. Assuming a linear model of identity and pose, we record the time series of the difference $dz_i$ between the latent code of each frame and the source's average latent vector. 

\begin{equation}
    \label{eq:source_latent}
    z^s_{id} = \frac{\sum z^s_i}{N} ~~~~~ dz_i = z^s_i - z^s_{id}
\end{equation}

We then transfer this latent pose to a still target image of Obama. There are two possibilities for obtaining the identity vector of the still target image $z^t_{id}$. One possibility is to select a canonical expressionless still image, and compute its latent code. This code is the approximate identity of the target person $z^t_{id}$. If there is a full video of the target person and our goal is to transfer the facial movement from the source person to the target person, we can compute the average identity vector (exactly as performed for the source person). Regardless of the method used to estimate the identity latent code of the target, reanimation is simply performed by adding the latent difference $dz_i$ to the target identity:  

\begin{equation}
    \label{eq:target_latent}
    z^t_i = z^t_{id} + dz_i
\end{equation}

To synthesize the animated target image, we pass each latent code through the generator $G$, to form the new video $\{G(z^t_i)\}$.

In practice we find that the codes computed by latent optimization of different video source frames contain differences in terms of brightness and the background. To reduce discontinuity artifacts, we filter the sequence of latent codes with a moving average. We find that this significantly smooths the synthesized motion.

Several frames of motion transfer performed by our method, are presented in Fig.~\ref{fig:animation}. We can observe that motion is transferred quite reliably between the source and target faces without using face specific supervision.

% One of the advantages of this approach is that it is not face specific. It can in fact be used for transferring motion between any video and still image, as long as they are from the same domain and that we have a trained  generator for this domain that can be inverted. In Fig.~\ref{} we present several frames from reanimation of a cat image, which simply uses a trained style-generative model of cat images (weights provided by the public GitHub repository of StyleGAN).

\section{Discussion}
\label{sec:disc}

\textit{Disentanglement by style architectures:} As an experiment, we trained GLO with the style architecture \cite{glo}, using a perceptual loss (as implemented in \cite{nam_eccv}). We found that the decompositional properties of the style (e.g. attribute flipping), are also present in the GLO results. We therefore believe that other properties such as reanimation could be performed in the future with non-adversarial architectures such as GLANN \cite{hoshen2018non}, which are easier and faster to train. 

\textit{Attribute flipping:} We found that using the codes obtained by latent optimization for each identity, we are able to perform attribute flipping exactly as for generated face presented in \cite{karras2018style}. This is remarkable, as it gives us an ability to flip attributes between real images without supervision. 

\textit{Per-layer latent codes:} Although StyleGAN is trained with a single latent code that is shared between all layers, we found that it is not sufficient for obtaining a good reconstruction of even in-distribution images. Instead, we performed latent optimization on the latent code input to each layer. This suggests that better generative models may be obtained by lifting the restriction of the global latent code entering all layers during StyleGAN training. We leave this investigation to future work.

\textit{Evaluation on LPIPS:} We elected to evaluate on LPIPS which is a perceptual similarity measure rather than $L_1$ or $L_2$ as our method goes through a much tighter bottleneck than DIP, which makes it easier for them to overfit to specific pixel values. To illustrate the superior quality of our results, we evaluate on a measure that emphasized perceptual similarity.

\textit{Invertible GAN architectures:} It has been supposed that due to the mode-dropping experienced in GAN training, the inversion of GAN trained generators will in general not be feasible. Several solutions were suggested (e.g. \cite{Peleg2018StructuredG}). We have shown that StyleGAN architectures are invertible. Our hypothesis is that the direct connection between the latent codes and the top network layers makes the inverse optimization much easier, and enables the invertibility. In the future, we intend to investigate alternative architectures for adding connections between the latent codes and the top layers of the network.

\textit{Feed-forward inference:} Latent optimization is effective and does not require training an encoder on a large number of images. The main disadvantage is that is slow at inference-time, as multiple feed-forward and backward steps through the network are required, as well as the evaluation of a perceptual loss. Although not essential to illustrating the main contributions made by this work, training a feed-forward encoder will make image reanimation less compute intensive and much faster. An analogous attempt for unsupervised domain translation can be seen in VAE-NAM \cite{vaenam}. An additional advantage of a feed-forward encoder was demonstrated by Zhu et al. \cite{zhu2016generative} that when starting from a latent code inferred by an encoder and then performing latent optimization yields better local optima than either latent optimization from random initialization or the results of encoder on its own.  

\section{Conclusions}
\label{sec:conc}

In this paper, we showed that the recently introduced StyleGAN generator architecture is invertible, differently from previous GAN-trained models. We argued that invertible generators are useful as strong image priors. The high quality of the image prior was clearly demonstrated on inpainting and super-resolution tasks. We also showed that the invertibility empowers us to use the linear decompositional properties of the generator latent space. This was used to transfer motion between a source video and a target still image.     

{\small
\bibliographystyle{ieee}
\bibliography{main}
}

\end{document}